\documentclass[10pt,twocolumn,letterpaper]{article}

\usepackage{iccv}
\usepackage{times}
\usepackage{epsfig}
\usepackage{graphicx}
\usepackage{amsmath}
\usepackage{amssymb}
\usepackage{geometry}
\usepackage{subcaption}
\usepackage{multirow}
\usepackage{xcolor}
\usepackage{enumitem}
\usepackage{bm}
\usepackage{color}
\usepackage{makecell}
\usepackage{algorithmic}
\usepackage{algorithm}
\usepackage[pagebackref=true,breaklinks=true,letterpaper=true,colorlinks,bookmarks=false,citecolor=blue,linkcolor=blue]{hyperref}
\newcommand{\eat}[1]{}

\newcommand\blfootnote[1]{%
	\begingroup 
	\renewcommand\thefootnote{}\footnote{#1}%
	\addtocounter{footnote}{-1}%
	\endgroup 
}
\usepackage[breaklinks=true,bookmarks=false]{hyperref}

\iccvfinalcopy 

\def\iccvPaperID{2560} 

\ificcvfinal\fi

\begin{document}

\title{Prototype-Aware Multimodal Alignment for Open-Vocabulary Visual Grounding}

\author{Jiangnan Xie$^\dagger$, Xiaolong Zheng$^\dagger$$^*$ Liang Zheng$^\dagger$\\
\small $^\dagger$ College of Electronics and Information, Hangzhou Dianzi University, Hangzhou, China\\
{\tt\small xlzheng@hdu.edu.cn}
}

\maketitle
\ificcvfinal\thispagestyle{empty}\fi

\begin{abstract}

Visual Grounding (VG) aims to utilize given natural language queries to locate specific target objects within images. 
While current transformer-based approaches demonstrate strong localization performance in standard scene 
(i.e, scenarios without any novel objects), 
they exhibit notable limitations in open-vocabulary scene (i.e, both familiar and novel object categories during testing). 
These limitations primarily stem from three key factors: (1) imperfect alignment between visual and linguistic modalities, 
(2) insufficient cross-modal feature fusion, and (3) ineffective utilization of semantic prototype information. 
To overcome these challenges, we present Prototype-Aware Multimodal Learning (PAML), an innovative framework that 
systematically addresses these issues through several key components: First, we leverage ALBEF to establish robust 
cross-modal alignment during initial feature encoding. 
Subsequently, our Visual Discriminative Feature Encoder selectively enhances salient object representations while 
suppressing irrelevant visual context. The framework then incorporates a novel prototype discovering and inheriting 
mechanism that extracts and aggregates multi-neighbor semantic prototypes to facilitate open-vocabulary recognition.
These enriched features undergo comprehensive multimodal integration through our Multi-stage Decoder before final 
bounding box regression. Extensive experiments across five benchmark datasets validate our approach, 
showing competitive performance in standard scene while achieving state-of-the-art results in open-vocabulary scene.
Our code is available at https://github.com/plankXie/PAML.
\end{abstract}

\blfootnote{*Corresponding Author: Xiaolong Zheng.} 

\vspace{-0.1in}
\section{Introduction}	
\label{sec:Introduction}
Visual Grounding (VG), also referred to as Phrase Grounding(PG)~\cite{hu2016natural,yang2020improving,he2024improved} or 
Referring Expression comprehension(REC)~\cite{yu2016modeling,mao2016generation,yu2017joint}, 
aims to accurately localize objects described in language expressions within corresponding images, 
based on a comprehensive understanding of the given image-language query pairs. Given its inherent requirement 
for deep comprehension of both visual and textual modalities, the effective resolution of VG tasks holds 
significant potential for application in various other multimodal domains, such as Vision-Language 
Navigation~\cite{anderson2018vision,yang20243d,qi2020reverie,zhan2023object}, Visual Question 
Answering~\cite{antol2015vqa,chen2020counterfactual,li2018visual,huang2019multi}, and Human-Robot 
Interaction~\cite{driess2023palm,rozenberszki2022language}.
However, most current models are trained using fully supervised learning, which requires access 
to annotated region bounding boxes and referring expressions. Manually annotating such data in 
large quantities is extremely costly. 
Therefore, in recent years, numerous methods have been proposed to alleviate this issue, including 
semi-supervised learning~\cite{zhu2021utilizing,jin2023pseudo,sun2023refteacher} (where some 
data is fully annotated and some is partially annotated), weakly supervised 
learning~\cite{sun2021cycle,wang2021weakly,chen2018knowledge,datta2019align2ground,wang2021improving} 
(where only images and corresponding text are provided, without bounding boxes), and unsupervised 
learning~\cite{xiao2023clip,yeh2018unsupervised,wang2019phrase,shi2023unpaired,jiang2022pseudo} (where only 
images are given, without any task-related annotations). 
Additionally, in the complex and diverse real-world scenarios, the objects to be located are often not 
necessarily encountered during the model's previous training. Consequently, addressing the model's performance 
in open-vocabulary scenes is more practically significant compared to standard scenes.

Open-vocabulary scenarios, where objects in the test set may not have appeared in the training set, 
are also referred to as general zero-shot settings in some literature~\cite{chen2021zero}. In recent years, 
large-scale vision-language pre-training models such as CLIP~\cite{radford2021learning}, ALBEF~\cite{li2021align}, 
VLMO~\cite{bao2022vlmo}, BLIP~\cite{li2022blip}, CoCa~\cite{yu2022coca}, and BEIT-3~\cite{wang2023image} have demonstrated 
robust feature representation capabilities, enabling seamless transfer to downstream 
tasks~\cite{tang2020blockmix,qian2023multimodal,tang2022learning,hu2021vivo,li2020oscar,gu2022openvocabulary}. 
It is natural for us to leverage these powerful models for open-vocabulary visual grounding tasks. 
However, directly applying these models without fully considering the intrinsic information within the 
image and text can be detrimental to object localization. The TransCP~\cite{tang2023context} paper innovatively 
proposes two methods: Context Disentangling and Prototype Discovering and Inheriting~\cite{li2020transferrable,xu2020attribute}, 
which significantly enhance the model's accuracy in both standard and open-vocabulary scenes. Nevertheless, this model 
only considers intrinsic contextual information and nearest prototype information, making it susceptible to discriminative 
relational information. 
When trained on datasets with low-quality context information and complex expressions, such as ReferIt, 
and tested on other datasets, the model's performance is suboptimal. Additionally, the modal 
fusion in this 
model relies solely on simple Hadamard fusion, and other similar Transformer-based methods also perform only superficial modality fusion through 
simple attention operations on dual-modal data. Therefore, more thorough modality fusion is critically necessary.

Based on a comprehensive analysis of the strengths and limitations of prior works, we propose a novel 
framework PAML designed to enhance model performance in both standard and open-vocabulary scenes. 
This framework incorporates the ALBEF module, leveraging its robust capabilities in multimodal representation and alignment 
to derive highly representative and well-aligned image and language features. The Visual Discriminative Feature Encoder can 
enhances discriminative representations of salient objects while suppressing irrelevant contextual information. Additionally, 
the framework integrates a Multiple Neighbor Prototype Discovering and Inheriting module, enabling the model to learn from a 
prototype bank and fully utilize the training data to ground referents in both scenes. 
Furthermore, a Multi-Stage Decoder module is employed to facilitate modality fusion, thereby promoting the model's localization 
capabilities across the two scenes.

\begin{figure*}[t] 
  \centering
  \includegraphics[width=0.9\textwidth]{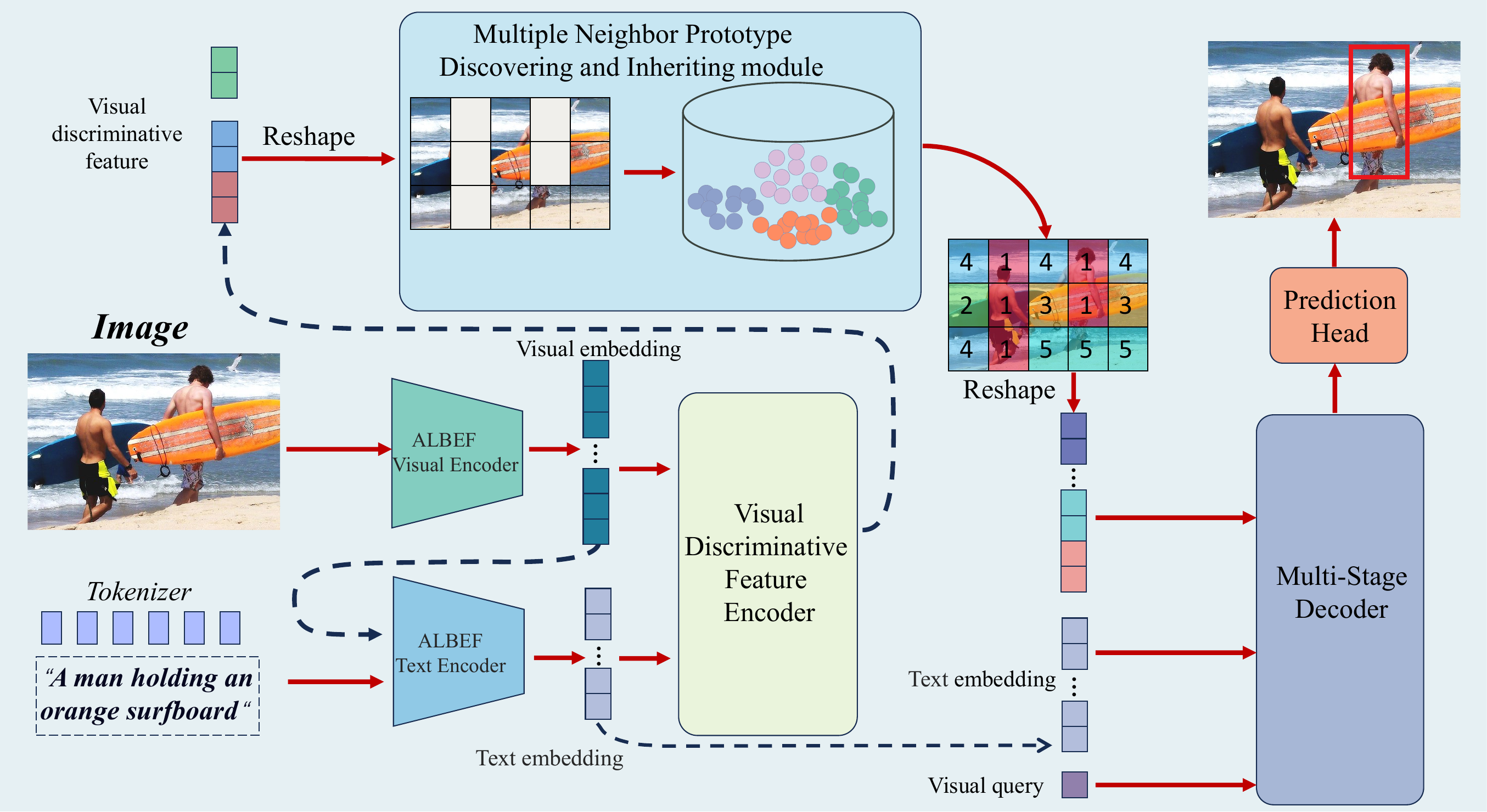} 
  \caption{Overall architecture of the proposed PAML. It first aligns image and text 
  features using ALBEF, then enhances object representations while suppressing irrelevant contextual 
  information through  Visual Discriminative Feature Encoder. Then it discovers and aggregates 
  multi-neighbor semantic prototypes for open-vocabulary recognition, before fusing multimodal features in a Multi-Stage Decoder. 
  Finally, the REG token is processed by the Prediction Head to regress the coordinates of the target bounding box.}
  \label{fig:framework}
\end{figure*}

As illustrated in Fig. \ref{fig:framework}, the image data is first processed by the ALBEF Visual Encoder to 
obtain its corresponding embeddings, while the textual data is tokenized and subsequently fed into the ALBEF 
Text Encoder with embeded image data for text embedding. These multimodal embeddings are then input into the Visual 
Discriminative Feature Encoder module to extract highly discriminative visual features. These features are subsequently 
fed into the prototype bank for the processes of discovering and inheriting, thereby acquiring multiple neighbor visual 
prototype features. The visual prototype features, along with the visual features unprocessed by the Visual Discriminative 
Feature Encoder module, are concatenated and combined with the visual query and the linguistic features derived from the 
ALBEF Text Encoder. This amalgamated data is then input into the Multi-Stage Decoder to facilitate modality fusion. 
Finally, the visual query is processed by the Prediction Head to obtain the bounding box (BBOX) coordinates of the target object.

In summary, we make five-fold contributions:
\begin{itemize}
	\item We proposed a novel and effective model, PAML, in which each module collaborates seamlessly. 
	 It demonstrates exceptional performance in both standard and open-vocabulary scenarios, achieving SOTA results across
	 multiple benchmark datasets. Extensive ablation experiments are conducted to validate the effectiveness of each individual 
	 component.
	\item We directly utilized a pre-trained ALBEF module for robust encoding and cross-modal alignment.
    \item Based on the visual context disentangling module in TransCP, we improved it into a Visual 
	Discriminative Feature Encoder by incorporating Laplacian transformation and balancing the contributions of 
	the two transformations through learnable parameters.
	\item We extend TransCP's Prototype Discovering and Inheriting by adopting multi-neighbor prototype 
	inheritance instead of nearest-neighbor inheritance, enhancing the richness of prototype representations.
    \item We proposed a novel Multi-Stage Decoder to address the issue of insufficient modality fusion 
	caused by direct attention operations on dual-modal data in previous Transformer-based methods.
\end{itemize}

\section{Related Work}
\label{sec:Related Work}
\subsection{Visual Grounding}
\label{sec:Visual Gounding}
Existing VG models can be broadly categorized into three types based on multimodal fusion and 
reasoning paradigms: two-stage methods, one-stage methods, and Transformer-based methods. 

Two-stage methods~\cite{hong2019learning,zhuang2018parallel,yang2019dynamic,yu2018mattnet,hu2017modeling,wang2019learning} 
rely on pre-trained object detectors to generate a set of candidate bounding boxes and then measure the similarity between 
these candidates and the language description to locate the target object. However, this approach is inherently constrained 
by the performance of the pre-trained detector. If the generated bounding boxes fail to encompass the target object requiring 
localization, the subsequent network is entirely incapable of accurately determining the object's position.
One-stage methods~\cite{yang2019fast,wang2022ofa,liao2022progressive,su2023language,yang2020improving,chen2018real,liao2020real} 
perform visual-language fusion after embedding the multimodal data and then output the bounding box with the highest score among 
predefined dense anchor points. Nevertheless, the intricate modality fusion modules, coupled with the anchor-based nature of this 
approach, impose significant limitations on the model's localization performance.

Transformer-based methods~\cite{vaswani2017attention,deng2021transvg,dai2024simvg,zheng2024resvg,Dai_Li_Zhuang_Zhang_Yang_2025,xiao2024hivg,xiao2023clip,tang2023context} 
are relatively more elegant. TransVG~\cite{deng2021transvg} first leverage Transformers to 
directly embed and fuse the two modal data, and utilizing a special REG token to directly regress 
the coordinates of the target object. This model demonstrates commendable performance and has been 
widely adopted as a baseline in numerous subsequent studies. Zheng et al.~\cite{zheng2024resvg} proposed the ResVG model, 
which leverages the Stable Diffusion model to generate a substantial amount of synthetic data, thereby enhancing the model's 
comprehension of fine-grained semantics and spatial relationships.
Dai et al.~\cite{dai2024simvg} introduced the SimVG model, which decouples visual-linguistic feature integration from 
downstream tasks by employing pre-trained multimodal models and supplementary object tokens. This model integrates a dynamic 
weight-balancing distillation strategy within a multi-path learning framework to reinforce a lightweight MLP pathway, thereby 
simplifying the architecture and significantly accelerating inference speed.
Xiao et al.~\cite{xiao2024hivg} introduced the HiVG model, which employs a cross-modal bridge to establish multi-level 
connections between visual and textual modalities. This approach addresses the inconsistency between visual features and the 
features required for grounding. Additionally, the model incorporates HiLoRA to adapt cross-modal features from shallow to deep 
layers, thereby preventing the accumulation of perceptual errors.
Transformer-based methods have demonstrated promising performance and are progressively emerging as mainstream approaches in the 
field of Visual Grounding tasks.

\subsection{Zero-Shot Visual Grounding}
\label{sec:Zero-Shot Visual Grounding}
As previously Sec.~\ref{sec:Introduction} mentioned, open-vocabulary is also referred to as general 
zero-shot settings. Therefore, research on zero-shot related work is enlightening for the study of open-vocabulary visual grounding. 
ZSGNet~\cite{sadhu2019zero} summarizes four scenarios for evaluating model performance and introduces a novel dataset for zero-shot 
visual grounding, establishing itself as a baseline for numerous subsequent studies. 
Shi et al.~\cite{shi2022improving} proposed leveraging external knowledge to construct a multimodal 
knowledge graph to address zero-shot visual grounding challenges. With the advancements in pre-trained vision–language models, 
a series of methods based on such models have emerged in recent years. 
For instance, Yao et al.~\cite{yao2024cpt} introduced CPT, which reformulates visual grounding as a fill-in-the-blank problem 
with co-referential markers based on color in both images and text, thereby minimizing the gap. The 
ReCLIP~\cite{subramanian-etal-2022-reclip} model employs isolated proposals to ingeniously transform the visual grounding 
problem into an image-text matching task, fully leveraging the capabilities of CLIP. Additionally, it incorporates a spatial 
relation resolver to address CLIP's limitations in spatial reasoning. The GroundVLP~\cite{shen2024groundvlp} model utilizes 
Grad-CAM to highlight regions highly relevant to the language query. Simultaneously, it employs an open-vocabulary object detector 
to identify all objects consistent with the subject in the expression, ultimately localizing the target object by integrating these 
two components.
The remarkable performance of pre-trained vision-language models in zero-shot visual grounding has inspired us to leverage models 
such as ALBEF and CLIP to enhance model capabilities in open-vocabulary scene.

\section{Methods}
\label{sec:Methods}
We propose PAML, a novel framework designed to effectively address 
the problem of open-vocabulary scene visual grounding. 
As illustrated in Fig. \ref{fig:framework}, the entire network is primarily composed of five key 
components: (1) ALBEF Encoder, (2) Visual Discriminative Feature Encoder, 
(3) Multiple Neighbor Prototype Discovering and Inheriting module, (4) Multi-Stage Decoder, and 
(5) Prediction Head. In the subsequent sections, we will elaborate on the design principles underlying 
each of these components in detail. The related concepts and definitions of $MHA(\cdot)$, $MCA(\cdot)$, 
and $MHARPE(\cdot)$ used in the equations can be found in Sec.~\ref{sec:Preliminary}.

\subsection{ALBEF Encoder}
\label{sec:ALBEF Encoder}
Given a preprocessed image $\mathit{I} \in \mathbb{R}^{B \times 3\times H_0\times W_0}(H_0=W_0=256)$ 
and a corresponding text $\mathit{T} \in \mathbb{R}^{B \times L_0}(L_0=40)$, the image data is initially 
fed into the ALBEF Visual Encoder. This module primarily adopts a ViT-B/16 architecture. The image undergoes 
patch embedding processing, transforming it into image tokens with a size of $\mathbb{R}^{B \times 256 \times C_0}$, $C_0$ 
is set to $768$. These tokens are then concatenated with a learnable class (cls) token $\in \mathbb{R}^{B \times 1\times C_0}$ 
initialized to zero. Subsequently, the combined tokens are processed through 12 layers of Transformer Encoder blocks. 
Following this, the tokens are expanded to 400 via linear interpolation to meet the dimensionality requirements of 
the subsequent network. Ultimately, image tokens  are obtained as the $f_v\in \mathbb{R}^{B \times 400\times C_0}$.

The input $\mathit{T}$ and $f_v$ are fed into the ALBEF text encoder. Initially, $\mathit{T}$ is passed through the 
BertEmbeddings module to obtain the preliminary text embeddings $f_{emb} \in \mathbb{R}^{B \times L_0\times C_0}$. 
Subsequently, $f_{emb}$ and $f_v$ are input into the BertEncoder module. Within each layer, self-attention operation is 
performed on the text features, followed by a cross-attention operation between the text and image data. 
This process is repeated across a total of 12 layers, ultimately yielding $f_l \in \mathbb{R}^{B \times L_0\times C_0}$ 
as the final output.
In our implementation, we directly employ an off-the-shelf version of ALBEF that has been pre-trained on 
the MSCOCO dataset. Since we employ the pre-trained ALBEF model, we are able to obtain aligned representations $f_l$ and $f_v$.

\subsection{Visual Discriminative Feature Encoder}
\label{sec:Visual Discriminative Feature Encoder}
Highlighting the features of salient objects in the image while suppressing irrelevant 
contextual information is crucial for subsequent prototype discovering and inheriting. 
Therefore, in the second step, $f_v$ and $f_l$ are fed into the visual discriminative feature 
encoder to obtain the discriminative image features.

Firstly, $f_v$ and $f_l$ undergo linear projection along the feature dimension, with the feature dimension 
size changing from $C_0$ to $C=256$. By utilizing Eq.~\eqref{eq:textinfo}, where $f_v$ is employed as the query 
and $f_l$ as both the key and value, a MCA operation is performed. This enables us to obtain the linguistic 
features $f_{linfo} \in \mathbb{R}^{400\times B \times C}$, which encapsulate information relevant to the objects within the image.

\begin{equation}
	\label{eq:textinfo}
	f_{linfo} = \text{MCA}(f_v, f_l, f_l)
\end{equation}

Then, a cosine similarity measurement operation is performed between $f_{linfo}$ and $f_v$ to 
obtain the similarity score $\phi_{sim} \in \mathbb{R}^{400\times B \times 1}$.
\begin{equation}
	\label{eq:sim}
	\phi_{sim} = \underset{i}{\sum}(\frac{f_{linfo(i)}}{\|\mathbf{f_{linfo}}\|_2} \cdot \frac{f_{v(i)}}{\|\mathbf{f_v}\|_2})
\end{equation}

where $i$ is the index over the feature dimensions of the data.
Subsequently, Gaussian transformation and Laplacian transformation are applied to $\phi_{sim}$ to enhance 
its robustness and expressive capability.

\begin{equation}
	\label{eq:gaussian}
	\phi_{G} = exp(-\frac{1-\phi_{sim}^2}{2\sigma^2})
\end{equation}

\begin{equation}
	\label{eq:Laplacian}
	\phi_{L} = exp(-\frac{|1-\phi_{sim}|}{b})
\end{equation}

The Gaussian transformation Eq.~\eqref{eq:gaussian} can smooth $\phi_{sim}$, making its 
data distribution more continuous and stable. This helps mitigate the impact of noise on 
similarity computation, thereby enhancing the robustness of the model. Additionally, the Gaussian transformation 
assigns higher weights to values close to $1$ (indicating high similarity) while suppressing values far from $1$, 
enabling the model to focus more on highly correlated image regions and text, thus improving the quality of feature fusion. 
On the other hand, the Laplacian transformation Eq.~\eqref{eq:Laplacian} models the absolute differences in $\phi_{sim}$, 
allowing it to better handle outliers or anomalies, which enhances the stability of the model when dealing with noisy or 
inaccurate similarity scores. Furthermore, the Laplacian transformation assigns certain weights to both high and low similarity 
regions, avoiding the potential issue of the Gaussian transformation overly emphasizing high-similarity areas. Finally, 
the data from both transformations are combined through a learnable parameter $\lambda$ (Eq.~\eqref{eq:phivisual}), which 
flexibly adjusts the weights of the two transformations to balance the emphasis on high-similarity regions and the robustness 
to outliers. Initially, $\lambda$ is set to $0.5$.

\begin{equation}
	\label{eq:phivisual}
	\phi_{v} = \lambda \phi_{G} + (1-\lambda)\phi_{L} 
\end{equation}
Then, similar to Eq.~\eqref{eq:textinfo}, we employ another MCA layer with different 
parameters to generate $f_{lcinfo} \in \mathbb{R}^{400\times B \times C}$, which represents 
the guiding information of the text for the image features. Subsequently, the image features are modulated 
according to Eq.~\eqref{eq:moduimg}. 

\begin{equation}
	\label{eq:moduimg}
	\begin{split}
	   f_{mv} = (f_{lcinfo} \times \alpha)\times f_{v} + f_{lcinfo} \times \beta 
    \end{split}	 
\end{equation}
where $\alpha \in \mathbb{R}^{C\times C}$ and  $\beta \in \mathbb{R}^{C\times C}$ are 
weight parameters, $f_{mv} \in \mathbb{R}^{400\times B \times C}$ and Eq.~\eqref{eq:moduimg} integrates 
the linguistic information into the image features.

\begin{align}
	\label{eq:textcimgctx}
	f_{lcv} = MHARPE(f_{mv}, f_{mv}, f_v)
\end{align}

Subsequently, we assign $q=k=f_{mv}$ and $v=f_v$  to execute the MHARPE operation, 
thereby deriving $f_{lcv} \in \mathbb{R}^{400\times B \times C}$. This procedure 
facilitates the incorporation of linguistic information into $f_{lcv}$, augmenting the 
representational capacity of the image features, while simultaneously enriching the 
contextual information of the image and capturing comprehensive global relationships.
Finally, Eq.~\eqref{eq:discriminative} is utilized to generate the discriminative image 
feature $f_{disv} \in \mathbb{R}^{400\times B \times C}$. 
Here, $Norm$ represents $LayerNorm$ with the parameter $\text{eps} = 1\text{e}-5$. we 
concatenate $f_v$ and $f_{disv}$ in feature dimension and output it to the following modules.

\begin{equation}
	\label{eq:discriminative}
	\begin{split}
		f_{disv} = (Norm(f_v) + Norm(f_{lcv})) \times \phi_{v}
	\end{split}
\end{equation}
While this module shares conceptual similarities with the visual 
context disentangling module in TransCP, a critical distinction lies in our dual-transformation approach. 
The TransCP implementation relies solely on Gaussian transformation for feature modulation, which may 
lead to excessive emphasis on high-similarity regions while aggressively suppressing low-similarity areas. 
To address this limitation, we introduce a complementary Laplacian transformation that provides more gradual 
weight distribution, coupled with a learnable parameter $\lambda$ to dynamically balance their contributions.
This synergistic combination, validated through ablation studies in Table~\ref{tab:ablation-transformation}, 
demonstrates superior robustness in preserving potentially useful weakly-correlated features while maintaining 
focus on strongly-aligned regions.

\subsection{Multiple Neighbor Prototype Discovering and Inheriting}
\label{sec:Multiple Neighbor Prototype Discovering and Inheriting}
Prototype information play a crucial role in open-vocabulary scene. However, the Prototype 
Discovering and Inheriting module in TransCP~\cite{tang2023context} only considers the nearest tokens 
to the input. In reality, other proximal prototypes could also contribute to localization to varying degrees. 
To address this limitation, we have enhanced that module to effectively capture neighboring prototype information 
within a certain distance threshold of the input token. The obtained $f_{disv}$ 
is initially upsampled to $768$ ($C_1=768$) dimensions via a 2D convolutional layer with both stride and kernel size 
set to $1$, 
which is denoted as $f_{in} \in \mathbb{R}^{B\times C_1 \times20\times20}$. Subsequently, 
the data from all batches are aggregated to form the $X \in \mathbb{R}^{(B\times400)\times C_1}$, 
which is then fed into the prototype bank module for the processes of discovering and inheriting. The 
corresponding pseudocode for this procedure is depicted in Algorithm \ref{alg:Algorithm of Prototype discovering and inheriting}.

\begin{algorithm}[!h]
    \caption{Algorithm of Multiple Neighbor Prototype Discovering and Inheriting}
    \label{alg:Algorithm of Prototype discovering and inheriting}
    \renewcommand{\algorithmicrequire}{\textbf{Input:}}
    \renewcommand{\algorithmicensure}{\textbf{Output:}}
    
    \begin{algorithmic}[1]
        \REQUIRE $X$, $E$, $S$, $C$, $k$    
        \ENSURE  $Q$ 
        
        \STATE  $d_{ij}={||X_i||^2 + ||E_j||^2 - 2\cdot X_i \cdot E_j^T}$
		\STATE  $N_i = topk(-d_{ij},k)$
		\STATE  $
					w_{ij} = 
					\begin{cases} 
					\frac{exp(-d_{ij}/\tau)}{\sum_{l\in N_i}exp(-d_{il}/\tau)} & \text{if } j = N_i \\
					0 & \text{otherwise}
					\end{cases}
					$
		\IF {$train$}
		\STATE $s_j = \sum_{i=1}^N w_{ij}$
		\STATE $S_j \leftarrow (1-\alpha)\cdot S_j + \alpha \cdot s_j$
		\STATE $c_j = w_{i,j}^T\cdot X_i$
		\STATE $C_j \leftarrow (1 - \alpha) \cdot C_j + \alpha \cdot c_j$
		\STATE $S \leftarrow \frac{S + \epsilon}{\sum_jS_j + n\cdot\epsilon}\times \sum_jS_j $
		\STATE $E_j \leftarrow \frac{C_j}{S_j}$
		\ENDIF
		\STATE $Q_i = \sum\limits_{j=1}^{k}w_{ij}Ej$
		\STATE $Q_i \leftarrow sg(Q_i-X_i)+X_i$		
        \RETURN Q
    \end{algorithmic}
\end{algorithm}

It is essential to initialize the prototype embedding $E \in \mathbb{R}^{2048\times C_1}$, the cluster 
size $S \in \mathbb{R}^{2048}$ for each token, and the mean value $C \in \mathbb{R}^{2048\times C_1}$ of the embeddings 
for each token within the prototype bank, all of which are set to zero.
The squared Euclidean distance between each element of $X_i$ and $E_j$ is computed, followed by obtaining the 
index $N \in \mathbb{R}^{B\times 400}$ of the nearest k prototype token based on this distance. Subsequently, 
the index $N$ is transformed into a weight matrix  $w \in \mathbb{R}^{(B\times 400)\times 2048}$. 
the temperature parameter $\tau$ is initialized to 1 and made learnable during training. 
If the process occurs during training, the prototype bank $E$ and its associated statistical quantities are updated. 
The update procedure begins by calculating the cluster size $s \in \mathbb{R}^{2048}$ for each prototype token $E_j$ 
(i.e., the number of input data points assigned to this prototype token). 
The cluster size $S \in \mathbb{R}^{2048}$ is then updated using Exponential Moving Average (EMA). 
Next, the embedded average $c \in \mathbb{R}^{2048\times C_1}$ for each prototype token $E_j$ 
(i.e., the sum of all input data points assigned to this prototype vector) is computed, 
and the token embedded average $C$ is updated using EMA. The decay rate $\alpha$ is set to $0.4$. 
To prevent the cluster size of certain prototype tokens from being zero, Laplace smoothing is applied to $S$, and $S$ 
is used to normalize the prototype tokens, thereby completing the update of $E$.

Subsequently, the input data is mapped to its nearest $k$ prototype token, yielding the 
quantized feature $Q \in \mathbb{R}^{(B\times 400)\times C_1}$. Finally, a stop-gradient operation is applied, 
ensuring that the quantized feature is utilized during forward propagation, while the original input is 
retained for backward propagation. Experimental results Sec.~\ref{sec:Ablation Study} demonstrate that 
the optimal performance is achieved when $k=5$.

After obtaining $Q$, it is reshaped into a tensor with the same spatial dimensions as the $f_{in}$, 
represented as $f_{qt}\in \mathbb{R}^{B\times C_1\times20\times20}$. Subsequently, it is concatenated with 
input along the feature dimension to form a tensor $T_{feat}$ with a size of $\mathbb{R}^{B\times(2\times C_1)\times20\times20}$.
According to Eq.~\eqref{eq:tscore}, dimensionality reduction and a softmax operation are applied to $T_{feat}$, yielding $T_{s}$.
The $Gate(\cdot)$ is implemented as a 2D convolutional layer with an input channel size of $1,536$, 
an output channel size of $2$, and both stride and kernel size set to $1$.
The data from the first feature dimension is extracted as $E_{s} \in \mathbb{R}^{B\times1\times20\times20}$, while 
the data from the second feature dimension is extracted as $I_{s} \in \mathbb{R}^{B\times1\times20\times20}$.
\begin{equation}
	\label{eq:tscore}
	T_{s} = Softmax(Gate(T_{feat}))
	\end{equation}

Finally, prototype out $P \in \mathbb{R}^{B\times C\times20\times20}$ will be obtained through following equation.

\begin{equation}
	\label{eq:proto}
	\begin{split}
		   P = Convt(f_{in}\times I_{s} + f_{qt}\times E_{s})
	\end{split}
\end{equation}

where the  $Convt$ is  a 2D convolutional layer with an input channel size of $768$, an output channel size of $256$, 
and both stride and kernel size set to $1$.	
The feature $f_{qt}$ captures the global structure and semantic information of the input data, 
effectively representing the distribution of the input data in a high-dimensional space, 
making it well-suited for tasks with significant inter-class variations. On the other hand, $f_{in}$ preserves 
the local details and low-level features of the input data, enabling it to capture fine-grained information. 
Through dynamic weighting mechanisms ($I_{s}$ and $E_{s}$), the model can adaptively integrate global semantic 
information with local detail information, thereby comprehensively extracting valuable insights from the 
data. $P$ will be reshaped to a size of $\mathbb{R}^{400\times B\times256}$, denoted as $f_q$.

\subsection{Multi-Stage Decoder}
\label{sec:Multi-Stage Decoder}
In this section, we delve into the intricate workings of the Multi-Stage Decoder. Previous research 
predominantly employed attention mechanisms to directly fuse modalities among REG tokens, visual tokens, 
and textual tokens. In contrast, our proposed methodology advocates for a multi-stage approach to modality fusion. 
The specific procedure is delineated as follows. Initially, a visual query token $f_{vq} \in \mathbb{R}^{1\times B\times C}$ is 
initialized as zero, and the $T_{info}$ is computed according to Eq.~\eqref{eq:Tinfo}. This involves extracting semantic 
information pertinent to the visual query from the textual features. Subsequently, the $V_{info}$ is derived using 
Eq.~\eqref{eq:Vinfo}, which entails extracting visual information relevant to the textual query from the image features. 
Finally, the corresponding visual query is obtained through Eq.~\eqref{eq:resfvq} and Eq.~\eqref{eq:resfvq2}.
The Feed-Forward Network (FFN) is composed of two Linear layers with a ReLU activation function in between. 
The intermediate dimension of the Linear layers is $2,048$, while the input and output dimensions of the FFN are both $256$.
\begin{equation}
	\label{eq:Tinfo}
	T_{info}^{n} = Norm(MCA(f_{vq}^{n-1},f_l,f_l))
\end{equation}

\begin{equation}
	\label{eq:Vinfo}
	V_{info}^{n} = Norm(MCA(T_{info}^{n},f_q,f_{v}))
\end{equation}

\begin{equation}
	\label{eq:resfvq}
	f_{vqt}^{n} = Norm(f_{vq}^{n-1}+V_{info}^{n})
\end{equation}

\begin{equation}
	\label{eq:resfvq2}
	f_{vq}^{n} = Norm(f_{vqt}^{n}+FFN(f_{vqt}^{n}))
\end{equation}

This module design exhibits substantial novelty compared to conventional approaches.
Compared to previous approaches in multimodal processing, the extraction of textual and visual information is 
conducted separately in a decoupled manner. This decoupled design enables the model to handle information from 
different modalities more flexibly, thereby avoiding direct interference between modalities. Simultaneously, the model 
leverages a nested attention mechanism, where textual information guides the extraction of visual information, 
and visual information, in turn, refines the representation of textual information. This nested architecture allows 
the model to more effectively capture the intricate interactions between modalities.

\subsection{Training Objective} 
\label{sec:Training Objective}
The Multi-Stage Decoder consists of a total of 6 layers, and we also utilize the intermediate-layer $f_{vq}^i$. 
The corresponding bounding box coordinates are obtained through the Prediction Head, as illustrated in Eq.~\eqref{eq:predictionhead}.

\begin{equation}
	\label{eq:predictionhead}
	\hat{B_i} = MLP(f_{vq}^i)
\end{equation}
The MLP consists of three Linear layers coupled with the Sigmoid activation function. The input and
output dimensions of the first two Linear layers are both $256$, while the output dimension of 
the final Linear layer is $4$, representing the four coordinates of the bounding box.
Following other Transformer-based methods~\cite{deng2021transvg}~\cite{tang2023context}, we adopt the $L1$ 
loss and the GIoU loss as the loss functions during the training phase,

\begin{equation}
	\label{func:loss}
	\mathcal{L} = \sum\limits_{i=1}^6(\lambda_{L1}\mathcal{L}_{L1}(B,\hat{B_i}) + \lambda_{GIoU}\mathcal{L}_{GIoU}(B,\hat{B_i})),
\end{equation}

\noindent where the factors $\lambda_{L1}$ and $\lambda_{GIoU}$ are set to 5 and 2 empirically to achieve an optimal balance between the two loss functions.
By incorporating the loss corresponding to intermediate queries, the model receives multi-stage supervision during the training process. This multi-stage supervision facilitates improved learning of localization tasks, mitigates the risk of overfitting, and enhances the model's generalization capabilities across diverse scenarios. Additionally, intermediate queries capture feature information at varying scales, which can assist the model in comprehensively understanding the context of the targets, thereby resulting in more accurate localization of those targets.

\subsection{Theoretical Analysis}
Our framework's effectiveness in open-vocabulary generalization is grounded in fundamental 
learning theory principles. Below, we provide a detailed theoretical justification for the some key innovations of our approach.
\subsubsection{Multi-Prototype Mechanism and Manifold Smoothing}
In open-vocabulary scenarios, novel objects often lie in low-density regions of the feature space, 
where training data is sparse. Previous methods relying on single-prototype matching (e.g., nearest-neighbor retrieval) 
suffer from high variance in these regions, leading to unstable predictions.
Our Multiple Neighbor Prototype Discovering and Inheriting module mitigates this issue by aggregating 
information from multiple semantically related prototypes. Given a prototype bank $\mathcal{P} = \left\{ p_{i} \right\}_{i=1}^{M}$ 
and an input feature $\phi(x)$ the inherited prototype feature is computed as:
\begin{equation}
    \label{eq:prototype-math}	
	\hat{\phi}(x)=\sum_{k\in\mathcal{N}_{K}(x)} w_{k}\cdot p_{k} \quad (w_{k}\propto\exp\left(-\left\|\phi(x)-p_{k}\right\|^{2} /\tau\right))
\end{equation}
\noindent where $\mathcal{N}_{K}(x)$ denotes the K-nearest prototypes, and $\tau$ is a temperature parameter controlling 
the sharpness of the weighting.
This formulation is motivated by the manifold hypothesis~\cite{belkin2004semi}, which posits that high-dimensional 
visual features of semantically similar objects lie near a low-dimensional manifold. When novel object features reside in 
low-density regions of the manifold (e.g., inter-class boundary zones), single-prototype matching forcibly assigns them to 
the nearest isolated prototype, thereby disrupting the local topological structure of the manifold. By interpolating between 
prototypes, our method effectively smooths the decision boundary in regions with limited training data. 
The generalization error $\epsilon$ for novel objects can be decomposed into bias and variance terms~\cite{ziegel2003elements}:
\begin{equation}
    \label{eq:error}	
    \mathcal{E} \leq \frac{C}{\sqrt{K}} + L \cdot \mathbb{E}\left[\min_{p \in \mathcal{P}} \|\phi(x) - p\|\right]
\end{equation}
\noindent where $C$ is a data-dependent constant, $L$ is the Lipschitz constant of the downstream predictor, 
The first term on the right-hand side represents the variance term, while the second corresponds to the bias 
term. As $K$ increases from 1, the variance term initially dominates - analogous to standard deviation decay 
in Monte Carlo sampling - causing the error to decrease at a rate of $\mathcal{O}(1/\sqrt{K})$. When $K$ becomes 
excessively large, the bias term dominates due to the introduction of irrelevant prototypes, leading to increased bias.
Our experiments Table~\ref{tab:ablation-prototype-number} show $K=5$ achieves optimal bias-variance trade-off.
\subsubsection{Multi-Stage Fusion as Iterative Refinement}
Inspired by the EM algorithm~\cite{dempster1977maximum}, the Multi-Stage Decoder can be decomposed into two steps: 
(1) Compute cross-modal attention to estimate the relevance between visual regions and language tokens; 
(2) Update the visual query $f_{vq}^n$ to maximize agreement between modalities.
Mathematically, at stage $n$, the update rule is:
\begin{equation}
	\label{eq:multi-stage}
	f_{vq}^{n} = \Gamma\left(f_{vq}^{n-1}, \mathcal{F}_{\mathrm{MCA}}\left(f_{vq}^{n-1}, f_{l}, f_{v}\right)\right)
\end{equation}
\noindent where $\Gamma$ is a nonlinear transformation. This iterative process progressively reduces the modal 
gap $\mathcal{D}_{\mathrm{KL}}\left(P_{\text{vis}} \| P_{\text{text}}\right)$
The use of intermediate supervision Equation\ref{func:loss}  introduces multiple gradient signals during training, 
which act as a form of implicit regularization. The total gradient is:
\begin{equation}
	\label{eq:loss-gradient}
	\nabla_{\theta}\mathcal{L}_{\text{total}} = \sum_{i=1}^{6}\lambda_{i}\nabla_{\theta}\mathcal{L}_{i} \approx \mathbb{E}\left[\nabla_{\theta}\mathcal{L}\right] + \mathcal{O}\left(\sigma/\sqrt{N}\right)
\end{equation}
where $\sigma$ is the gradient noise. By averaging over multiple stages, 
the variance of the gradient estimate is reduced by a factor of $\sqrt{6}$, making optimization more stable 
and less prone to overfitting.
\subsection{Synergy between modules}
The PAML framework's effectiveness stems from a tightly-coupled pipeline where each component's output 
strategically feeds into and enhances the next. The ALBEF Encoder's cross-modal aligned representations 
serve as the critical foundation, enabling the Visual Discriminative Feature Encoder to perform targeted 
suppression of irrelevant visual context—this purification process directly optimizes the input for the 
subsequent prototype discovery stage. The Multiple Neighbor Prototype Discovering and Inheriting module 
then leverages these noise-reduced features to construct discriminative semantic clusters, with its 
multi-prototype aggregation mechanism dynamically guided by the linguistic embeddings from ALBEF. 
These enriched prototypes and the original visual features form complementary streams that the Multi-Stage 
Decoder intelligently fuses: at each decoding layer, textual semantics steer the attention to relevant prototype features, 
while the prototypes conversely disambiguate linguistic references through visual evidence. This bidirectional, 
hierarchical interaction creates a positive feedback loop—where better alignment enables more precise discrimination, 
which yields higher-quality prototypes, ultimately leading to more accurate fusion and regression. The modules' interdependence 
transforms them from isolated processors into an integrated reasoning system that progressively bridges visual 
and linguistic domains.

\section{Experiments}
\label{sec:Experiments}

\subsection{Datasets}
\label{sec:Datasets}
\noindent{\bf ReferIt.} The ReferIt dataset~\cite{kazemzadeh2014referitgame}, also referred to 
as ReferItGame or ReferCLEF, comprises approximately 20,000 annotated images sourced from the 
SAIAPR-12~\cite{escalante2010segmented} collection. Each image is associated with one or multiple 
region-specific referring expressions, though the dataset includes some ambiguous queries (e.g., "any," "whole") 
and occasional labeling errors. Following standard protocols in prior research, the dataset is partitioned into 
three subsets: a training set with $54,127$ expressions, a validation set containing 5,842 expressions, and a test set with $60,103$ 
expressions. The validation set is typically utilized for experimental analysis, while the test set serves as the 
benchmark for method comparison.

\noindent{\bf Flickr30K Entities.} Flickr30K Entities~\cite{plummer2017flickr30k} augments the Flickr30K~\cite{young2014image} 
dataset with short phrase-to-region annotations, comprising 31,783 images annotated with five referring expressions 
per image, yielding ~427K localized phrase-region pairs. The dataset focuses on succinct referential phrases 
rather than lengthy descriptions and excludes images containing multiple instances of the same object category. 
Following standard protocols, the data is partitioned into 29,783 training, 1,000 validation, and $1,000$ test images.

\noindent{\bf RefCOCO/ RefCOCO+/ RefCOCOg.} Derived from MSCOCO~\cite{lin2014microsoft}, 
the RefCOCO family comprises three benchmark datasets for referring expression comprehension. RefCOCO~\cite{yu2016modeling} 
contains 19,994 images with $50,000$ annotated objects and $142,209$ expressions, partitioned into standard 
splits: training ($120,624$), validation ($10,834$), and two test sets - testA ($5,657$, person-centric) and testB ($5,095$, 
object-focused). Its extension, RefCOCO+\cite{yu2016modeling}, maintains comparable scale ($19,992$ images, $49,856$ objects, $141,564$ 
expressions) but prohibits spatial descriptors, emphasizing appearance attributes instead. The dataset is officially 
partitioned into training, validation, testA, and testB subsets containing $120,191$, $10,758$, $5,726$, and $4,889$ 
expressions correspondingly. RefCOCOg~\cite{mao2016generation} expands the collection with $25,799$ images and $95,010$ expressions, 
featuring two evaluation protocols: the google split~\cite{mao2016generation} (used 
in our experiments for consistency with prior work) and the umd split~\cite{nagaraja2016modeling}.

\begin{table*}[t!]
	\caption{Comparisons of standard scene with state-of-the-art methods on RefCOCO~\cite{yu2016modeling}, RefCOCO+~\cite{yu2016modeling} and RefCOCOg~\cite{mao2016generation} in terms of top-1 accuracy (\%). We highlight the best and second best performance in the \textbf{bold} and \underline{underline}.}
	
	\vspace{-0.2cm}
	
	\small
	\begin{center}
		\scalebox{0.9}[0.9]{
			\setlength
			\tabcolsep{9.4pt}
			\begin{tabular}{c | c | c | c c c | c c c | c}
				\hline
				\multirow{2}{*}{Models} & \multirow{2}{*}{Venue} & \multirow{2}{*}{Backbone} & \multicolumn{3}{c|}{RefCOCO} & \multicolumn{3}{c|}{RefCOCO+} & \multicolumn{1}{c}{RefCOCOg} \\ 
				
				&  &  & val & testA & testB & val & testA & testB & val-g\\
				\hline \hline
				\textbf{\textit{Two-stage:}} & &  & & & & & & &\\
				CMN~\cite{hu2017modeling} & \textit{CVPR'17} & VGG16 & - & 71.03 & 65.77 & - & 54.32 & 47.76 & 57.47 \\
				VC~\cite{zhang2018grounding} & \textit{CVPR'18} & VGG16 & - & 73.33 & 67.44 & - & 58.40 & 53.18 &62.30 \\
				ParalAttn~\cite{zhuang2018parallel} & \textit{CVPR'18} & VGG16 & - & 75.31 & 65.52 & - & 61.34 & 50.86 & 58.03 \\
				A-ATT~\cite{deng2018visual} & \textit{CVPR'18} & VGG16 & - & 80.87 & 71.55 & - & 65.13 & 55.01 & 63.84 \\
				MAttNet~\cite{yu2018mattnet} & \textit{CVPR'18} &	ResNet-101 & 76.65 & 81.14 & 69.99 & 65.33 & 71.62 & 56.02 & - \\
				LGRANs~\cite{wang2019neighbourhood} & \textit{CVPR'19} & VGG16 & - & 76.60 & 66.40 & - & 64.00 & 53.40 & 61.78 \\
				DGA~\cite{yang2019dynamic} & \textit{ICCV'19} & VGG16 & - & 78.42 & 65.53 & - & 69.07 & 51.99 & -\\ 
				RvG-Tree~\cite{hong2019learning} & \textit{TPAMI'19} & ResNet-101 & 75.06 & 78.61 & 69.85 & 63.51 & 67.45 & 56.66 & - \\
				NMTree~\cite{liu2019learning} & \textit{ICCV'19} & ResNet-101 & 76.41 & 81.21 & 70.09 & 66.46 & 72.02 & 57.52 & 64.62 \\
				Ref-NMS~\cite{chen2021ref} & \textit{AAAI'21} & ResNet-101 & 80.70 & 84.00 & 76.04 & 68.25 & 73.68 & 59.42 & - \\
				\hline
				\textbf{\textit{One-stage:}} & &  & & & & & & &\\
				SSG~\cite{chen2018real} & \textit{arXiv'18} &  DarkNet-53 & - & 76.51 & 67.50 & - & 62.14 & 49.27 & 47.47 \\  
				FAOA~\cite{yang2019fast} & \textit{ICCV'19} & DarkNet-53 & 72.54 & 74.35 & 68.50 & 56.81 & 60.23 & 49.60 & 56.12 \\
				RCCF~\cite{liao2020real} & \textit{CVPR'20} & DLA-34 & - & 81.06 & 71.85 & - & 70.35 & 56.32 & -  \\
				ReSC-Large~\cite{yang2020improving} & \textit{ECCV'20} & DarkNet-53 & 77.63 & 80.45 & 72.30 & 63.59 & 68.36 & 56.81 & 63.12 \\
				LBYL-Net~\cite{huang2021look} & \textit{CVPR'21} & DarkNet-53 & 79.74 & 82.71 & 73.84 & 68.96 & 73.52 & 59.44 & 63.78 \\
				\hline
				\textbf{\textit{Transformer-based:}} & &  & & & & & & &\\
				RefTR~\cite{li2021referring}& \textit{NeurIPS'21} & ResNet-50 & 80.92 & 83.40 & 75.78 & 69.22 & 74.61 & 60.95 & 61.76 \\
				TransVG~\cite{deng2021transvg} & \textit{ICCV'21} & ResNet-50 & 80.48 & 82.57 & 75.79 & 66.22 & 71.25  & 57.39 & 67.41 \\
				Pseudo-Q~\cite{jiang2022pseudo} & \textit{CVPR'22} & ResNet-50 & 56.02 & 58.25 & 54.13 & 38.88 & 45.06 & 32.13 & 49.82 \\
				VGTR~\cite{du2022visual}  & \textit{ICME'22} & ResNet-50 & 80.48 & 82.52 & 75.79 & 63.46 & 70.21 & 53.57 & 63.76 \\
				SeqTR~\cite{zhu2022seqtr}  & \textit{ECCV'22} & DN53 & 78.22 & 81.47 & 73.80 & 66.01 & 70.23 & 55.68 & 68.26\\
				VLTVG~\cite{yang2022improving} & \textit{CVPR'22} & ResNet-50 & 83.21 & 86.78 & 78.45 & 72.36 & 77.21 & 64.80 & 71.40\\
				CLIP-VG~\cite{xiao2023clip} & \textit{TMM'23} & CLIP-B & 84.29 & 87.76 & 78.43 & 69.55 & 77.33 & 57.62 & 72.64\\		 		
				TransVG++~\cite{deng2023transvg++}	 & \textit{TPAMI'23} & ViT-Det & \textbf{86.28} & 88.37 & \underline{80.97} & \textbf{75.39} & \textbf{80.45} & \textbf{66.28} & 73.86\\
				ScanFormer~\cite{su2024scanformer} & \textit{CVPR'24} & ViLT & 83.40 & 85.86 & 78.81 & 72.96 & 77.57 & 62.50 & 74.10\\
				TransCP~\cite{tang2023context} & \textit{TPAMI'24} & ResNet-50 & 84.25 & 87.38 & 79.78 & 73.07 & 78.05 & 63.35 & 72.60\\
				ResVG~\cite{zheng2024resvg} & \textit{ACM MM'24} & ResNet-50 & 85.51 & \textbf{88.76} & 79.93 & 73.95 & \underline{79.53} & 64.88 & 73.13\\
				MGCross~\cite{miao2023self} & \textit{TIP'24} & ResNet-101 & 85.10 & \underline{88.23} & 80.08 & \underline{74.44} & 79.48 & \underline{65.21} & \underline{74.50} \\
				PAML(Ours) & - & ViT-B/16 & \underline{85.68} & 88.07 & \textbf{83.47} & 72.97 & 76.70 & 61.79 & \textbf{74.54}\\
				\hline
			\end{tabular}
		} 
	\end{center}
	\label{tab:refcoco_results}
	\vspace{-0.5cm}
\end{table*}

\subsection{Implementation Details}
\noindent{\textbf{Inputs.}} The input images are resized to $256\times256$ resolution, where the longer 
edge will be resized and the shorter edge will be padded. For referring expressions, we maintain a 
fixed sequence length of $40$ tokens, including special tokens [CLS] and [SEP]. Expressions exceeding $38$ 
tokens are truncated, while shorter sequences are padded with empty tokens after the [SEP] 
token to maintain consistent dimensionality.

\noindent{\textbf{Metric.}} In line with established evaluation protocols~\cite{deng2021transvg,jiang2022pseudo,li2021referring}, 
we employ top-1 localization accuracy as our primary metric. A predicted bounding box is deemed correct when its 
intersection-over-union (IoU) with the ground truth exceeds the conventional threshold of $0.5$.

\noindent{\textbf{Training Details.}} We employ the AdamW optimizer to train the model, 
with the initial learning rate set to $1e-4$. The ALBEF model, which has a size of $209.5$ million parameters, 
is pretrained on the MSCOCO dataset. The learning rates for the image encoder branch and the text encoder branch of ALBEF are set to 
$1e-5$. The learnable parameters $\sigma$ and $b$ are initialized to $0.5$ and $1$, respectively. 
The $\lambda$ parameter, which balances the contributions of terms $\phi_{G}$ and $\phi_{L}$, is initialized 
to $0.5$. All other learnable parameters are initialized using Xavier initialization. The model is trained for 
$90$ epochs across all datasets, with the learning rate decaying to 1e-5 after $60$ epochs.
according to~\cite{yang2022improving}, we set $\lambda_{L1}$ and $\lambda_{GIoU}$ to $5$ and $2$ respectively.

\noindent{\textbf{Inference.}} Following~\cite{tang2023context}, the model is initially trained using standard 
supervised learning protocols on established benchmarks, with model selection determined by hightest validation 
accuracy. The comprehensive testing procedure encompasses two critical aspects: conventional performance evaluation 
on standard test sets followed by an analysis of open-vocabulary generalization capabilities. 
To effectively assess the model's ability to handle novel vocabulary, we implement a systematic cross-dataset 
evaluation strategy where models trained on ReferIt are tested on RefCOCO series and Flickr30k Entities, those 
trained on RefCOCO are evaluated on ReferIt and Flickr30k, and models developed on Flickr30k are assessed on RefCOCO 
series and ReferIt. This carefully designed evaluation paradigm perfectly aligns with the fundamental 
requirements of open-vocabulary scenarios by necessitating generalization to completely unseen datasets 
containing both familiar and novel object categories.

All model training and inference procedures were executed on a single NVIDIA GeForce RTX 4090.

\subsection{Comparisons with State-of-the-art Methods}
\label{sec:Comparisons with State-of-the-art Methods}
\subsubsection{Comparisons methods}
\label{sec:Comparisons methods}
As described in the Related Work, the comparison methods
in the experiments can be categorized into three groups:
\begin{itemize}
\item Two-stage methods: CMN~\cite{hu2017modeling}, VC~\cite{zhang2018grounding}, 
ParalAttn~\cite{zhuang2018parallel}, A-ATT~\cite{deng2018visual}, MAttNet~\cite{yu2018mattnet}, 
LGRANs~\cite{wang2019neighbourhood}, DGA~\cite{yang2019dynamic}, RvG-Tree~\cite{hong2019learning}, 
NMTree~\cite{liu2019learning}, Ref-NMS~\cite{chen2021ref}, Similarity Net~\cite{wang2019learning}, CITE~\cite{plummerCITE2018}, 
PIRC~\cite{kovvuri2018pirc}, and DDPN~\cite{yu2018rethinking}. 
\item One-stage methods: SSG~\cite{chen2018real}, FAOA~\cite{yang2019fast}, RCCF~\cite{liao2020real}, 
ReSC-Large~\cite{yang2020improving}, LBYL-Net~\cite{huang2021look}, and ZSGNet~\cite{sadhu2019zero}.
\item Transformer-based methods: RefTR~\cite{li2021referring}, TransVG~\cite{deng2021transvg}, 
Pseudo-Q~\cite{jiang2022pseudo}, VGTR~\cite{du2022visual} , SeqTR~\cite{zhu2022seqtr}, 
VLTVG~\cite{yang2022improving}, CLIP-VG~\cite{xiao2023clip}, TransVG++~\cite{deng2023transvg++}, 
ScanFormer~\cite{su2024scanformer}, TransCP~\cite{tang2023context}, ResVG~\cite{zheng2024resvg}, and MGCross~\cite{miao2023self}.  
\end{itemize}

\begin{table}[t!]
	\caption{Comparisons of standard scene with state-of-the-art methods on the test set of  ReferIt~\cite{kazemzadeh2014referitgame} and Flickr30K Entities~\cite{plummer2017flickr30k} in terms of top-1 accuracy ($\%$). We highlight the best and second best performance in the \textbf{bold} and \underline{underline}.}
	\small
	\begin{center}
		\scalebox{0.9}{
			\begin{tabular}{c | c | c | c  }
				\hline
				\multirow{2}{*}{Models} & \multirow{2}{*}{Backbone} & \multicolumn{1}{c|}{ReferIt} & \multicolumn{1}{c}{Flickr30K}\\
				& & test & test\\
				\hline
				\hline
				\textbf{\textit{Two-stage:}}  &  &  &\\
				CMN~\cite{hu2017modeling} & VGG16 & 28.33 & -\\
				VC~\cite{zhang2018grounding} & VGG16 & 31.13 & - \\
				MAttNet~\cite{yu2018mattnet} &	ResNet-101  & 29.04 & - \\
				Similarity Net~\cite{wang2019learning} & ResNet-101 & 34.54 & 60.89\\
				CITE~\cite{plummerCITE2018}  & ResNet-101  & 35.07 & 61.33\\
				PIRC~\cite{kovvuri2018pirc} & ResNet-101 & 59.13 & 72.83 \\
				DDPN~\cite{yu2018rethinking} & ResNet-101 &  63.00 & 73.30\\
				\hline
				\textbf{\textit{One-stage:}}  &  & &\\
				SSG~\cite{chen2018real} & DarkNet-53 & 54.24 & - \\
				ZSGNet~\cite{sadhu2019zero} & ResNet-50 & 58.63 & 63.39\\
				FAOA~\cite{yang2019fast} & DarkNet-53 & 60.67 & 68.71\\
				RCCF~\cite{liao2020real} & DLA-34 & 63.79 & -\\
				ReSC-Large~\cite{yang2020improving} & DarkNet-53 & 64.60 & 69.28\\
				LBYL~\cite{huang2021look} & DarkNet-53 & 68.14 & -\\
				\hline
				\textbf{\textit{Transformer-based:}}  &  &   &\\
				Pseudo-Q~\cite{jiang2022pseudo} & ResNet-50 & 43.32 & 60.41\\
				RefTR~\cite{li2021referring} & ResNet-50 & 69.04 & 74.75\\
				VGTR~\cite{du2022visual} & ResNet-50 & 63.63 & 75.44\\
				SeqTR~\cite{zhu2022seqtr} & DarkNet-53 & 66.97 & 76.26\\
				VLTVG~\cite{yang2022improving} 	& ResNet-50 & 71.41 & 78.62\\
				TransVG~\cite{deng2021transvg} & ResNet-50 & 69.86 & 78.07\\
				CLIP-VG~\cite{xiao2023clip} & CLIP-B & 70.89 & \textbf{81.99}\\
				TransVG++~\cite{deng2023transvg++} & ViT-Det & 73.17 & \underline{81.49}\\
				ScanFormer~\cite{su2024scanformer} & ViLT & 68.85 & -\\
				TransCP~\cite{tang2023context} & ResNet-50 & 72.05 & 80.04\\
				ResVG~\cite{zheng2024resvg} & ResNet-50 & 72.35 & 79.52\\
				MGCross~\cite{miao2023self} & ResNet-101 & \underline{75.18} & -\\
				PAML(Ours) & ViT-B/16 & \textbf{75.65} & 79.51\\
				\hline
			\end{tabular}
		} 
	\end{center}
	\label{tab:referit_results}
	\vspace{-0.10cm}
\end{table}

\begin{figure*}[t!] 
   \centering
   \includegraphics[width=0.9\textwidth]{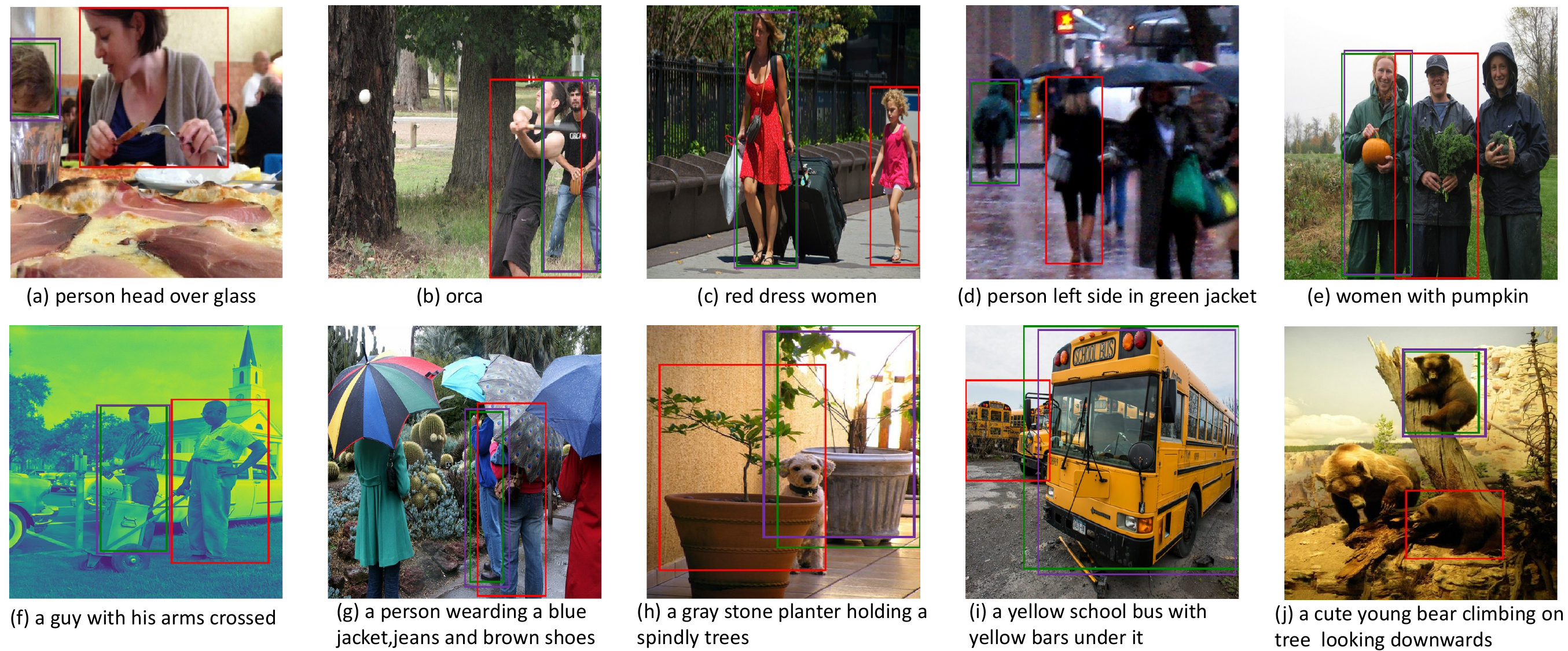} 
   \caption{Qualitative results of our model compared with the TransCP in the standard scene: trained on RefCOCO training set and tested on RefCOCO testA set(first row), and trained on RefCOCOg training set and tested on RefCOCOg val set (bottom row). Green: ground truth; Red: TransCP; Purple: PAML.}
    \label{fig:standard-result}
 \end{figure*}

\subsubsection{Comparisons in the Standard Scene}
\label{sec:Comparisons in the Standard Scene}
From Table~\ref{tab:refcoco_results}, which presents the performance of different 
models on the RefCOCO series datasets (with the highest accuracy highlighted in \textbf{bold} and the 
second-highest \underline{underlined}), it can be observed that our method demonstrates highly 
competitive results compared to state-of-the-art (SOTA) approaches. On the RefCOCO validation set, 
our method achieves an accuracy of $85.68\%$, which approaches the performance of the best-performing method, 
TransVG++ $86.28\%$. Notably, our method achieves an accuracy of $83.47\%$ on the RefCOCO testB dataset, surpassing the 
second-best method by $2.5\%$. 
The performance advantage on testB is particularly significant as it demonstrates our model's robustness in 
grounding less common or more varied object types compared to person-centric instances. Since the testB dataset 
exclusively contains non-person objects, this result strongly indicates the superior capability of our model in handling 
diverse object categories.
On the RefCOCOg val-google dataset, our method also achieves state-of-the-art performance with an accuracy 
of $74.54\%$. Given that referring expressions in RefCOCOg are generally more verbose and linguistically 
complex compared to other benchmarks, this result demonstrates our model's superior capability in 
processing long-form textual descriptions.
Transformer-based methods generally outperform both Two-stage and One-stage approaches. For instance, 
on the RefCOCO dataset, the best-performing Two-stage method, Ref-NMS, demonstrates 
accuracy deficits of $4.98\%$, $4.07\%$, and $7.43\%$ 
compared to our method on the val, testA, and testB subsets, respectively. This indicates that 
Two-stage methods suffer from significant limitations due to the potential exclusion of target objects during 
the proposal generation phase. 
Similarly, on the RefCOCO+ dataset, the top One-stage approach, LBYL-Net, exhibits accuracy gaps 
of $4.01\%$, $3.18\%$, and $2.35\%$ relative to our method. These results suggest that One-stage methods are 
constrained by their reliance on predefined dense anchor points.

\begin{table*}[!t]
	\caption{Comparison of open-vocabulary scene with the state-of-the-art methods by the 
	models trained on ReferIt and test on RefCOCO/+/g and Flickr30K Entities in terms of top-1 accuracy ($\%$). 
	We highlight the best and second best performance in the \textbf{bold} and \underline{underline}.}
	
	\vspace{-0.2cm}
	
	\small
	\begin{center}
		\scalebox{1}[1]{
			\setlength
			\tabcolsep{9.4pt}
			\begin{tabular}{c | c c | c c c | c c c | c}
				\hline
				\multirow{2}{*}{Models} & \multicolumn{2}{c|}{Flickr30k} & \multicolumn{3}{c|}{RefCOCO} & \multicolumn{3}{c|}{RefCOCO+} & \multicolumn{1}{c}{RefCOCOg} \\ 
				
				&  val & test & val & testA & testB & val & testA & testB & val-g\\
				\hline 
				\hline
				LBYL-Net~\cite{huang2021look} & 26.00 & 26.19 & 55.61 & 61.75 & 46.26 & 38.03 & 43.14 & 29.29 & 40.02  \\
				RefTR~\cite{li2021referring} & 29.88 & 30.47 &  58.16 & 61.15 & 52.83 & 36.49 & 40.15 & 32.67 & 44.96 \\
				VGTR~\cite{du2022visual} & 14.31 & 15.01 & 10.03 & 7.25 & 12.91 & 11.1 & 7.75 & 13.60 & 8.44 \\
				SeqTR~\cite{zhu2022seqtr} & 17.64 & 18.38 & 9.96 & 9.61 & 10.13 & 11.05 & 9.78 & 11.82 & 5.76 \\
				VLTVG~\cite{yang2022improving} & 51.40 & 53.13 & \underline{64.54} & \underline{65.83} & \textbf{61.98} & \underline{41.04} & \underline{43.70} & \textbf{38.33} & 47.11 \\
				TransVG~\cite{deng2021transvg} & 52.29 & 54.38 & 61.67 & 63.23 & 59.39 & 37.52 & 39.13 & 35.10 & 45.79\\
				TransCP~\cite{tang2023context} & \underline{52.92} & \underline{55.07} & 64.26 & 65.55 & \underline{61.71} & 40.17 &  42.70 & 36.90 & \underline{47.40}\\
				ResVG~\cite{zheng2024resvg} & 52.67 & 54.71 & 64.02 & 65.58 & 59.99 & 39.67 & 41.32 & 37.03 & 46.92\\
				PAML(Ours) & \textbf{53.33} & \textbf{55.63} & \textbf{65.04} & \textbf{67.16} & 60.1 & \textbf{44.03} & \textbf{47.43}  & \underline{37.96}  & \textbf{48.89} \\
				\hline
			\end{tabular}
		} 
	\end{center}
	\label{tab:open-referit}
	\vspace{-0.5cm}
\end{table*}

\begin{table}[t!]
	\caption{Comparison of open-vocabulary scene with the state-of-the-art methods by the models 
	trained on RefCOCO dataset and test on ReferIt and Flickr30K Entities in terms of top-1 accuracy ($\%$). 
	We highlight the best and second best performance in the \textbf{bold} and \underline{underline}.}
	\small
	\begin{center}
		\scalebox{0.9}{
			\begin{tabular}{c | c c | c c }
				\hline
				\multirow{2}{*}{Models} & \multicolumn{2}{c|}{ReferIt} & \multicolumn{2}{c}{Flickr30K}\\
				& val & test & val & test\\
				\hline
				\hline
				LBYL-Net~\cite{huang2021look} & 21.76 & 22.93 & 22.97 & 21.62 \\
				RefTR~\cite{li2021referring} & 27.67 & 29.45 & 24.03 & 23.08 \\
				VGTR~\cite{du2022visual} & 23.11 & 23.46 & 9.57 & 9.17 \\
				SeqTR~\cite{zhu2022seqtr} & 28.78 & 29.53 & 15.45 & 13.97 \\
				VLTVG~\cite{yang2022improving} & \underline{38.76} & 40.54 & 25.28 & 24.31 \\
				TransVG~\cite{deng2021transvg} & 36.11 & 37.86 & 23.78 & 22.83 \\
				TransCP~\cite{tang2023context} & 38.35 & \underline{40.62} & \underline{29.01} & \underline{27.71} \\
				ResVG~\cite{zheng2024resvg} & 37.62 & 39.54 & 25.77 & 26.58 \\
				PAML(Ours) & \textbf{41.70} & \textbf{42.78} & \textbf{29.07} & \textbf{27.75}\\
				\hline
			\end{tabular}
		} 
	\end{center}
	\label{tab:open-refcoco}
	\vspace{-0.10cm}
\end{table}

In Table~\ref{tab:referit_results}, our method achieves state-of-the-art performance on the ReferIt 
dataset with an accuracy of $75.65\%$. The ReferIt dataset presents significant challenges due to 
its inherent characteristics: multiple referring expressions may describe the same target 
object (e.g., "red cup" versus "coffee cup"), requiring the model to comprehend synonyms and coreference 
relationships. Furthermore, the natural language descriptions frequently contain ambiguous expressions (e.g., "that thing"), 
necessitating effective disambiguation through contextual understanding and visual grounding.
The superior performance of our model on this challenging benchmark demonstrates its robustness in handling 
linguistic variability and referential ambiguity.

To more intuitively demonstrate the advantages of our model, we compare its performance with the 
TransCP model under standard scene settings. As shown in Fig. \ref{fig:standard-result}
(b), (d), (e), and (i), PAML exhibits superior localization capability for small objects. For instance, 
in Fig. \ref{fig:standard-result} (i), where TransCP fails to localize correctly, PAML accurately identifies 
the target bus location by leveraging small visual cues ("yellow bars"). Similarly, in Fig. \ref{fig:standard-result} (b), 
PAML precisely localizes the target based on the small logo ("orca") on clothing.
Fig. \ref{fig:standard-result} (c), (f), (g), and (h) further illustrate PAML's ability to localize objects 
by effectively utilizing attribute-level descriptions. For example, in Fig. \ref{fig:standard-result} (f), PAML correctly 
identifies the target person with "arms crossed" against a greenish background. In Fig. \ref{fig:standard-result} (g), 
it successfully localizes the subject matching the textual description ("blue jacket," "jeans," and "brown shoes").
Additionally, Fig. \ref{fig:standard-result} (a) and (j) highlight PAML's robust spatial understanding. Specifically, 
in Fig. \ref{fig:standard-result} (a), PAML accurately localizes the target head position by reasoning about 
the relative position of the "glass", whereas TransCP fails under the same condition.

\subsubsection{Comparisons in the Open-Vocabulary Scene}
\label{sec:Comparisons in the Open-Vocabulary Scene}

\begin{figure*}[t!] 
   \centering
   \includegraphics[width=0.9\textwidth]{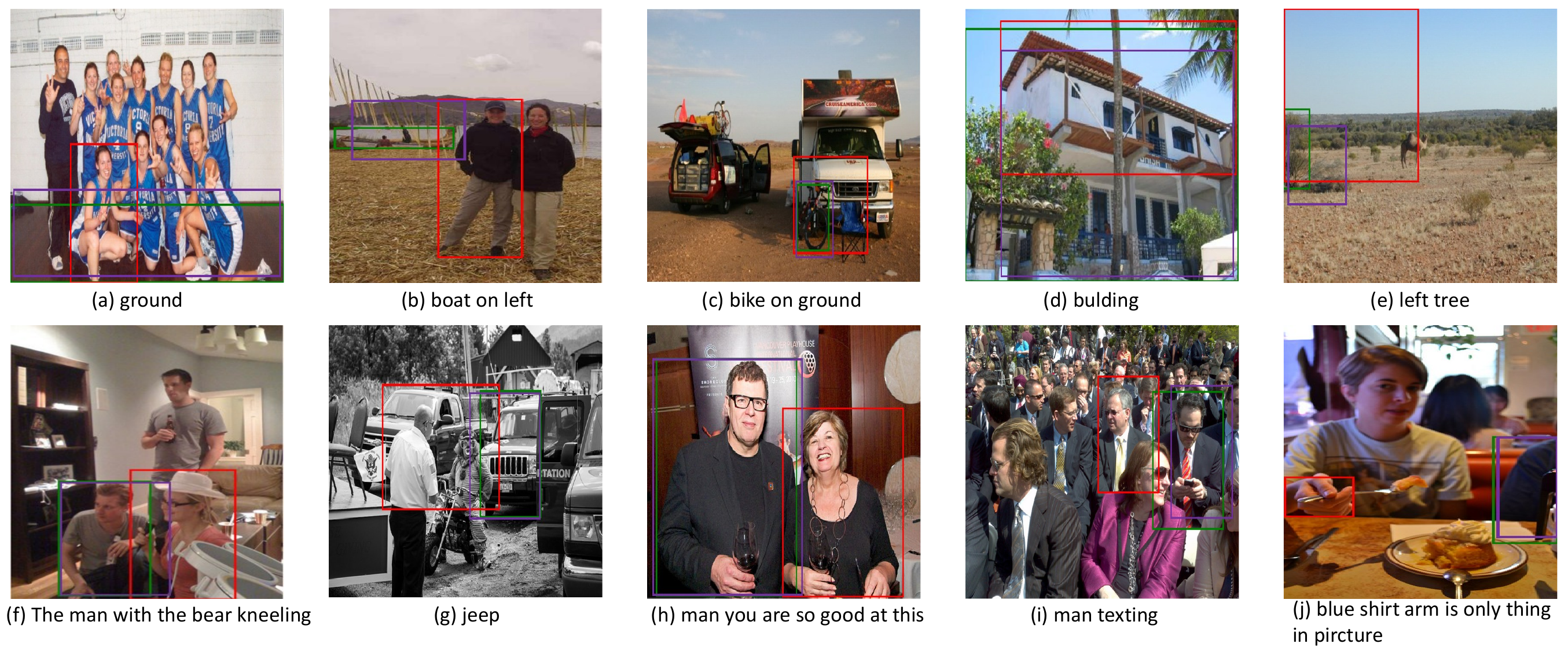} 
   \caption{Qualitative results of our model compared with the TransCP in the open-vocabulary scene: trained on RefCOCO training set and tested on ReferIt test set(first row), and trained on ReferIt training set and tested on RefCOCO+ testA set (bottom row). Green: ground truth; Red: TransCP; Purple: PAML.}
    \label{fig:open-result}
 \end{figure*}

\begin{table*}[t]
	\caption{Comparison of open-vocabulary scene with the state-of-the-art methods by the 
	models trained on Flickr30K dataset and test on RefCOCO/+/g and ReferIt in terms of top-1 accuracy ($\%$). 
	We highlight the best and second best performance in the \textbf{bold} and \underline{underline}.}
	
	\vspace{-0.2cm}
	
	\small
	\begin{center}
		\scalebox{1}[1]{
			\setlength
			\tabcolsep{9.4pt}
			\begin{tabular}{c | c c | c c c | c c c | c}
				\hline
				\multirow{2}{*}{Models} & \multicolumn{2}{c|}{ReferIt} & \multicolumn{3}{c|}{RefCOCO} & \multicolumn{3}{c|}{RefCOCO+} & \multicolumn{1}{c}{RefCOCOg} \\ 
				
				&  val & test & val & testA & testB & val & testA & testB & val-g\\
				\hline 
				\hline
				RefTR~\cite{li2021referring} & 36.61 & 35.70 & 29.72 & 36.16 & 23.21 & 27.67 & 31.85 & 22.21 & 35.13 \\
				VGTR~\cite{du2022visual} & 8.59 & 8.49 & 18.33 & 17.35 & 11.85 & 9.09 & 5.78 & 11.95 & 13.05 \\
				SeqTR~\cite{zhu2022seqtr} & 11.80 & 11.83 & 12.41 & 12.52 & 11.38 & 12.70 & 12.96 & 11.63 & 18.44 \\
				VLTVG~\cite{yang2022improving} & 44.13 & 42.25 & 39.67 & \textbf{46.33} & \textbf{31.82} & 36.28 & 41.83 & \underline{31.34} & 38.63\\
				TransVG~\cite{deng2021transvg} & 43.07 & 41.08 & 36.82 & 44.67 & 30.70 & 35.17 & 41.97 & 31.01 & \underline{44.85}\\
				TransCP~\cite{tang2023context} & \underline{45.36} & \textbf{43.99} & \textbf{39.35} & 46.26 & \underline{31.76} & \textbf{37.80} & \textbf{42.91} & \textbf{32.95} & \textbf{44.91}\\
				ResVG~\cite{zheng2024resvg} & 44.96 & 42.58 & 37.01 & 45.86 & 30.32 & 36.71 & 41.83 & 30.91 & 44.62\\
				PAML(Ours) & \textbf{45.64} & \underline{43.48} & \underline{38.73} & \underline{46.30} & 29.87 & \underline{36.82} & \underline{42.54}  & 30.70  & 42.09 \\
				\hline
			\end{tabular}
		} 
	\end{center}
	\label{tab:open-flickr}
	\vspace{-0.5cm}
\end{table*}

Based on the Sec.~\ref{sec:Comparisons in the Standard Scene}, it can be concluded that 
Transformer-based methods exhibit superior performance compared to both Two-stage and 
One-stage approaches. Therefore, for the sake of conciseness, this chapter will exclusively 
focus on comparative analyses among Transformer-based methods
LBYL-Net~\cite{huang2021look}, RefTR~\cite{li2021referring}, VGTR~\cite{du2022visual}, 
SeqTR~\cite{zhu2022seqtr}, VLTVG~\cite{yang2022improving} , TransVG~\cite{deng2021transvg}, TransCP~\cite{tang2023context}, 
and ResVG~\cite{zheng2024resvg}.
We conduct experiments under three settings:
\begin{itemize}
	\item Train on RefCOCO, test on ReferIt and Flickr30K.
	\item Train on ReferIt, test on RefCOCO series and Flickr30K.
	\item Train on Flickr30K, test on RefCOCO series and ReferIt.
\end{itemize}
As shown in Table ~\ref{tab:open-refcoco}, the proposed approach
 yields state-of-the-art performance in this experimental configuration, 
 with accuracy scores of $29.07\%$ (val) and $27.75\%$ (test) on Flickr30K, demonstrating a 
 slight but consistent improvement over TransCP(baseline). our method achieves an accuracy of $41.70\%$ on the ReferIt 
 validation set and $42.78\%$ on the test set, surpassing the second-best approach by $2.84\%$ and $2.16\%$, respectively. 
 These results indicate that the model possesses robust cross-dataset generalization ability, enabling it to effectively 
 generalize features from the 80 object categories in COCO and adapt them to other datasets with high accuracy.

Table ~\ref{tab:open-referit} demonstrates that our method delivers exceptional performance in this evaluation setting, 
achieving state-of-the-art results on all benchmark datasets with the sole exception of the RefCOCO/RefCOCO+ testB subsets. 
Particularly impressive are the model's accuracies of $44.03\%$ (val) and $47.43\%$ (testA) on RefCOCO+, which exceed those of the 
closest competitor by $2.99\%$ and $3.73\%$ respectively.
As noted in the Sec.~\ref{sec:Datasets}, the training data in RefeIt dataset comprises substantial amounts of complex and 
noisy annotations, such as ambiguous referring expressions and mis-labeled regions. These inherent data imperfections present 
considerable learning difficulties. Remarkably, our approach still establishes new state-of-the-art results, providing compelling 
evidence for its robustness in real-world scenarios characterized by complex linguistic expressions and imperfect visual grounding 
annotations.

From Table ~\ref{tab:open-flickr}, it can be observed that under this experimental setup, our method still maintains 
strong performance and achieves the second-highest accuracy on most datasets. For example, on the RefCOCO/RefCOCO+ validation 
and testA sets, our method attains accuracies of $38.73\%$, $46.30\%$, $36.82\%$, and $42.54\%$, respectively, trailing the 
top-performing method by only $0.62\%$, $0.03\%$, $0.98\%$, and $0.37\%$.
However, we note that on the RefCOCO/RefCOCO+ testB sets, our method achieves lower accuracies of $29.87\%$ and $30.70\%$. 
This performance gap may stem from the fact that the Flickr30K training dataset primarily contains person-centric and 
common object categories, whereas testB includes more rare or sparse object categories. Consequently, the Multiple Neighbor 
Prototype Discovering and Inheriting module may fail to learn representative prototype features for these objects, 
leading to suboptimal localization performance.

For a more intuitive demonstration of our model's advantages, we compare its performance with the baseline 
TransCP model under open-vocabulary settings in Fig. \ref{fig:open-result}.
Fig. \ref{fig:open-result} (a), (d), and (g) demonstrate PAML's robust localization capability even when provided with 
only a single-word query. This scenario requires precise and comprehensive semantic understanding of the isolated term. 
For instance, in Fig. \ref{fig:open-result} (d), the model accurately localizes the "building" while producing more reasonable 
results than the ground truth annotation.
Fig. \ref{fig:open-result} (e), (b), and (j) highlight PAML's superior performance in small-object localization. Notably, in 
Fig. \ref{fig:open-result} (j), the model successfully localizes the challenging target "blue shirt arm" where TransCP 
completely fails. Furthermore, 
Fig. \ref{fig:open-result} (i) showcases PAML's exceptional ability to interpret and localize human actions, 
as evidenced by its accurate detection of the "texting" behavior.

\begin{table*}[t]
	\caption{Ablation study of different modules in standard scene.}
	
	\vspace{0.1cm}
	
	\small
	\begin{center}
		\scalebox{0.9}[0.9]{
			\setlength
			\tabcolsep{9.4pt}
			\begin{tabular}{c c c c | c | c c c | c c c}
				\hline
				\multicolumn{4}{c|}{Modules} & \multicolumn{1}{c|}{ReferIt} & \multicolumn{3}{c|}{RefCOCO} & \multicolumn{3}{c|}{RefCOCO+}\\ 
				
				   AL   &  VD & PT & MD  & test & val & testA & testB & val & testA & testB\\
				\hline
				\hline
				\checkmark &		   & 		     & 		      & 72.52 & 80.99 & 82.97 & 76.32 & 66.71 & 72.25 & 57.49\\
				\checkmark &\checkmark &            &             & 73.42 & 82.73 & 85.07 & 77.23 & 68.42 & 74.85 & 58.31\\
				\checkmark &           & \checkmark & 			  & 72.61 & 81.02 & 83.02 & 76.41 & 66.93 & 72.32 & 57.53\\
				\checkmark &	       &            & \checkmark  & 74.53 & 84.09 & 86.02 & 81.98 & 70.99 & 75.45 & 60.09\\
				\checkmark &	       & \checkmark & \checkmark  & 74.87 & 84.42 & 86.75 & 82.27 & 71.33 & 75.78 & 60.21\\
				\checkmark &\checkmark & 		    & \checkmark  & 75.49 & 85.51 & 87.86 & 83.19 & 72.68 & 76.64 & 61.53\\
				\checkmark &\checkmark & \checkmark & 		      & 74.72 & 85.01 & 86.21 & 83.02 & 72.13 & 76.28 & 61.44\\
				           &\checkmark & \checkmark & \checkmark  & 73.91 & 84.93 & 86.63 & 82.87 & 71.84 & 75.73 & 60.97\\
				\checkmark &\checkmark & \checkmark & \checkmark  & 75.65 & 85.68 & 88.07 & 83.47 & 72.97 & 76.70 & 61.79\\

				\hline
			\end{tabular}
		} 
	\end{center}
	\label{tab:ablation-components-standard}
	\vspace{-0.5cm}
\end{table*}

\begin{table}[t]
	\caption{Ablation study of different modules in open-vocabulary scene, trained on RefCOCO and test on ReferIt.}
	\small
	\begin{center}
		\scalebox{1}{
			\begin{tabular}{c c c| c c}
				\hline
				\multicolumn{3}{c|}{Modules} & \multicolumn{2}{c|}{ReferIt}\\
				VD & PT & MD  & val & test\\
				\hline
				\hline
						   & 		    & 		      & 36.71 & 38.44 \\
				\checkmark &            &             & 37.42 & 39.21 \\
				           & \checkmark & 			  & 36.94 & 38.51 \\
					       &            & \checkmark  & 38.04 & 39.55 \\
					       & \checkmark & \checkmark  & 38.51 & 40.07 \\
				\checkmark & 		    & \checkmark  & 40.28 & 40.67 \\
				\checkmark & \checkmark & 		      & 41.31 & 42.29 \\
				\checkmark & \checkmark & \checkmark  & 41.70 & 42.78 \\

				\hline
			\end{tabular}
		} 
	\end{center}
	\label{tab:ablation-component-open}
	\vspace{-0.10cm}
\end{table}

\subsubsection{Ablation Study}
\label{sec:Ablation Study}

In this section, we will present a comprehensive validation of the contribution of each 
module within the model to the overall experimental outcomes.
Table ~\ref{tab:ablation-components-standard} and Table ~\ref{tab:ablation-component-open} present 
the ablation study results in the standard-scene and open-vocabulary scene, respectively.
AL, VD, PT, and MD denote the ALBEF Encoder, Visual Discriminative Feature Encoder, Multiple Neighbor 
Prototype Discovering and Inheriting module, and Multi-Stage Decoder, respectively. A checkmark (\checkmark) 
indicates module usage, while a blank entry denotes its exclusion. Without employing ALBEF, we substitute both 
the image encoder and text encoder with pre-trained models of comparable scale: ViT-B/16 (85.8M parameters) 
for visual processing and BERT-base (123.7M parameters) for textual processing. Generally, the experimental 
results demonstrate that the model's accuracy declines when any individual module is omitted. 

In Table ~\ref{tab:ablation-components-standard}, we observe that employing only the Multi-Stage Decoder (MD) 
yields the most significant performance improvement in the standard-scene compared to using either of the other two 
modules alone. For instance, on the RefCOCO val, testA, and testB datasets, adding only MD improves performance by $3.1\%$, $3.05\%$, 
and $5.66\%$, respectively, compared to the baseline without any of these modules.

The performance improvement can be attributed to our Multi-Stage Decoder architecture, which effectively captures 
fine-grained semantic relationships between extended linguistic expressions and visual features through iterative cross-modal fusion.
Furthermore, the results indicate that employing the Multiple Neighbor Prototype Discovering and Inheriting module 
(PT) alone yields marginal performance gains. For instance, on the RefCOCO+ dataset, integrating only PT improves 
accuracy by merely 0.23\%, 0.07\%, and 0.04\% across different splits compared to the baseline configuration. 
However, a substantial performance boost is observed when PT is combined with the Visual Discriminative Feature 
Encoder (VD). On the ReferIt test set, the joint use of VD and PT achieves a $2.2\%$ accuracy improvement, 
significantly surpassing the individual contributions of VD ($0.9\%$) and PT ($0.09\%$). This suggests that the 
prototype discovery and inheritance mechanism effectively functions when paired with a Visual Discriminative 
feature encoder that can first highlight target objects while suppressing irrelevant contextual information.
Furthermore, a performance gap of $1.74\%$ in accuracy is also observed between ALBEF and the baseline model 
using separate ViT-B/16 and BERT-base encoders on the ReferIt test dataset. It can be primarily attributed to 
ALBEF's superior multimodal fusion architecture and training paradigm, which enables more effective cross-modal 
alignment through its integrated cross-attention mechanisms and contrastive learning objectives, as opposed to 
the late fusion approach of separately pre-trained unimodal encoders that may suffer from representation incompatibility 
and insufficient fine-grained visual-textual correspondence optimization.

\begin{table}[t]
	\caption{Ablation study of different transformations in standard scene.}
	\small
	\begin{center}
		\scalebox{1}{
			\begin{tabular}{c c c| c c c}
				\hline
				\multicolumn{3}{c|}{Modules} & \multicolumn{3}{c|}{RefCOCO}\\
				gaussian & laplacian & $\lambda$ & val & testA & testB\\
				\hline
				\hline
						   & 			& 			 & 84.91 & 87.51 & 82.64 \\
				\checkmark & 			& 			 & 85.27 & 87.73 & 82.84 \\
						   & \checkmark & 			 & 85.21 & 87.81 & 82.90 \\
				\checkmark & \checkmark &  			 & 85.32 & 87.90 & 83.29 \\
				\checkmark & \checkmark & \checkmark & 85.68 & 88.07 & 83.47 \\ 

				\hline
			\end{tabular}
		} 
	\end{center}
	\label{tab:ablation-transformation}
	\vspace{-0.10cm}
\end{table}

Similar observations can be drawn from Table~\ref{tab:ablation-component-open}. Notably, 
on the ReferIt test set, the VD+PT module combination achieves 1.03\% higher accuracy than VD+MD. However, 
the opposite trend is observed in Table ~\ref{tab:ablation-components-standard} (standard-scene setting). 
This indicates that the PT module effectively captures and transfers object-level prototype information across datasets, 
thereby enhancing localization performance. In contrast, since standard scenes do not involve cross-category 
discrepancies between training and test sets, the performance gain from prototype transfer is less pronounced.

\begin{figure*}[t] 
	\centering
	\begin{subfigure}[b]{0.195\linewidth}
		\includegraphics[width=\textwidth]{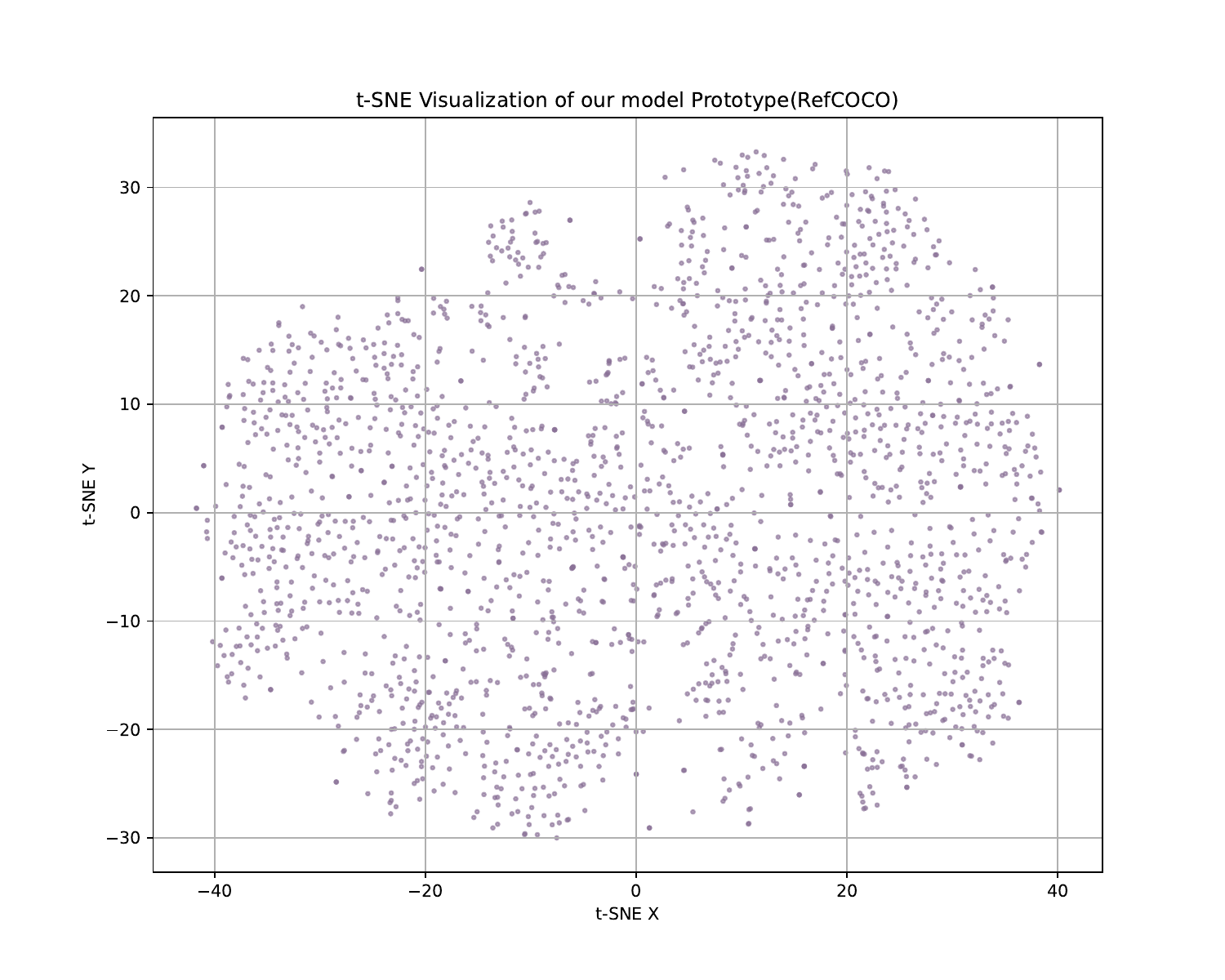}
		\caption{PAML: RefCOCO}
		\label{fig:our_refcoco}
	\end{subfigure}
	\hfill
	\begin{subfigure}[b]{0.195\linewidth}
		\includegraphics[width=\textwidth]{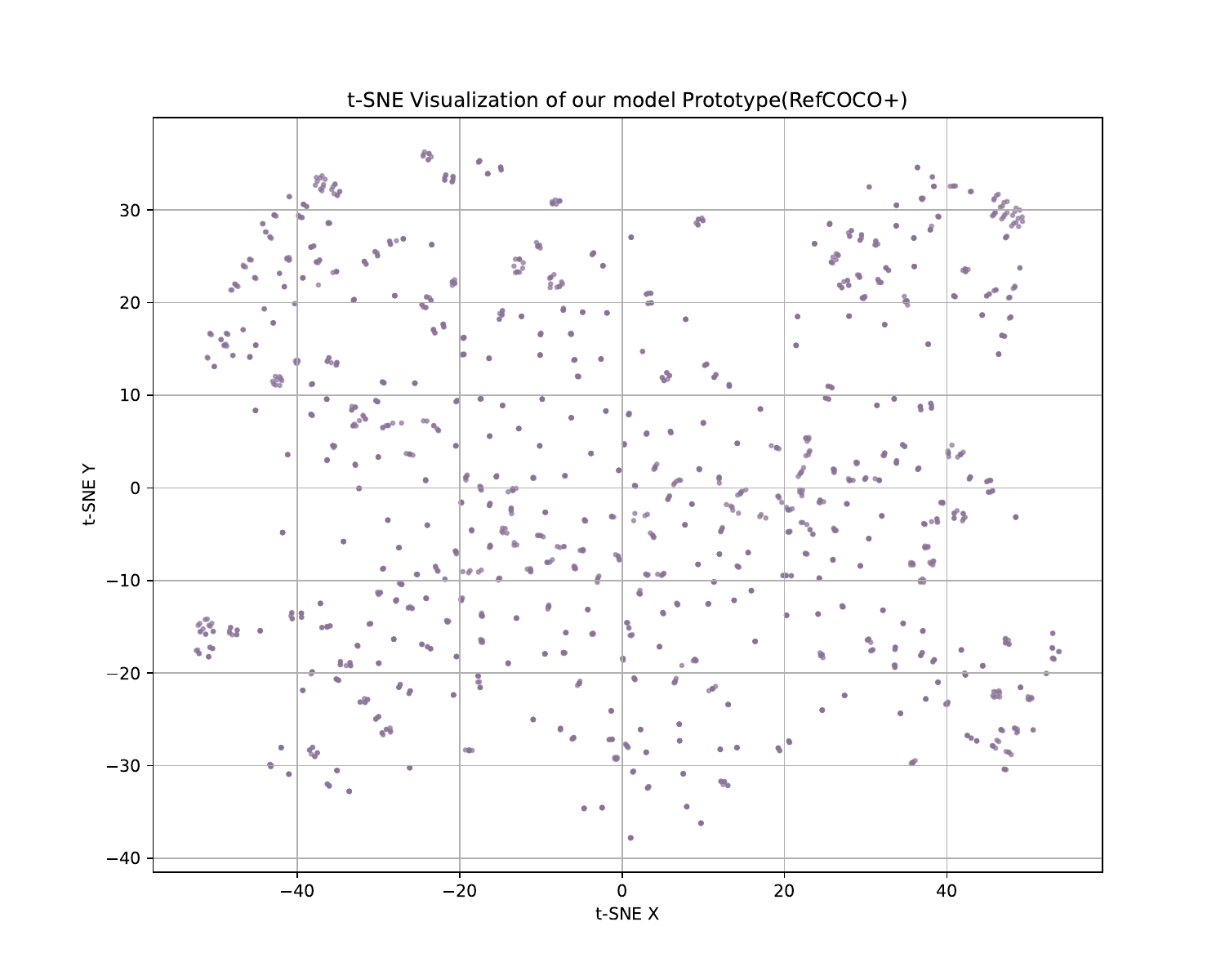}
		\caption{PAML: RefCOCO+}
		\label{fig:our_refcocoplus}
	\end{subfigure}
	\hfill
	\begin{subfigure}[b]{0.195\linewidth}
		\includegraphics[width=\textwidth]{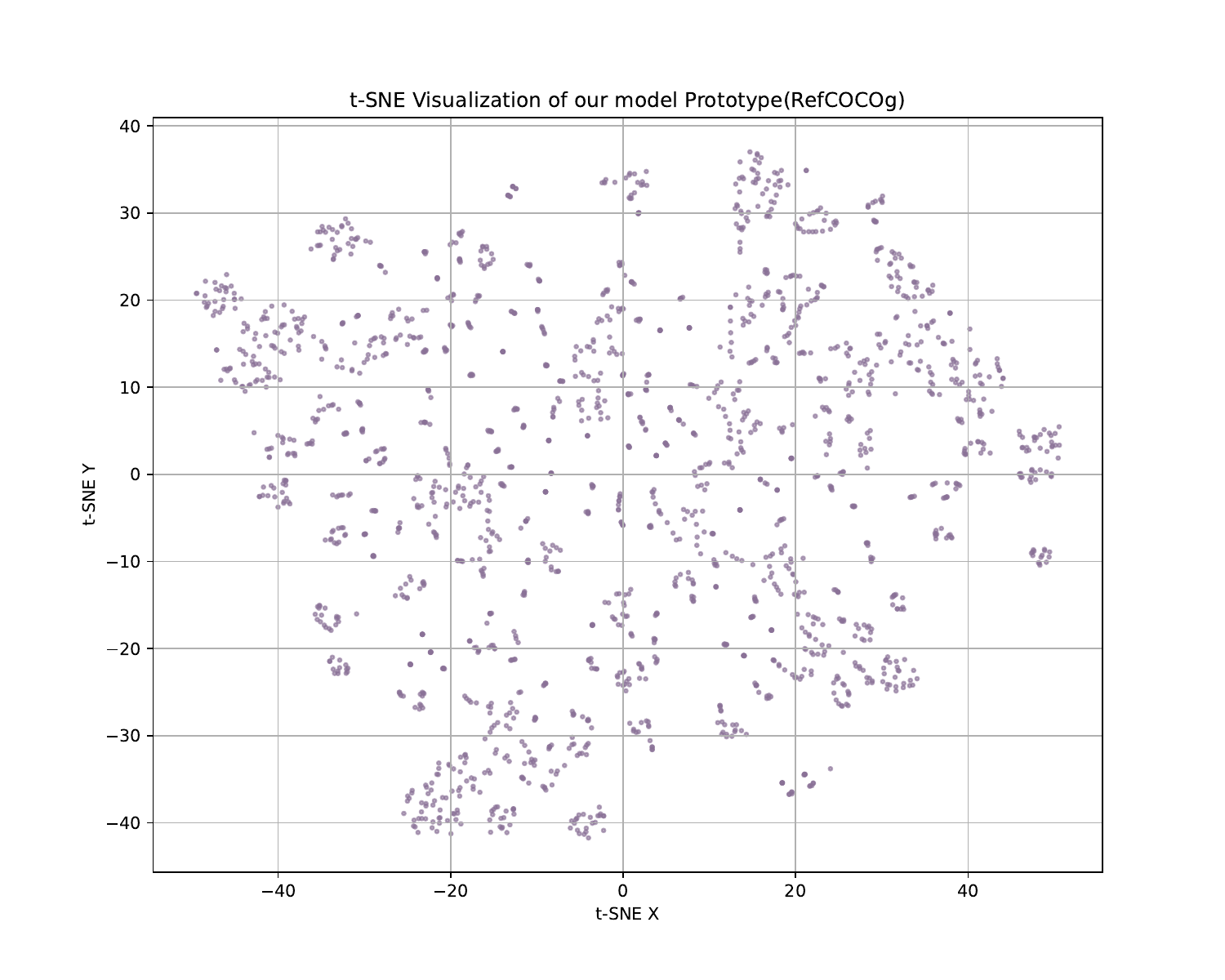}
		\caption{PAML: RefCOCOg}
		\label{fig:our_refcocog}
	\end{subfigure}
	\hfill
	\begin{subfigure}[b]{0.195\linewidth}
		\includegraphics[width=\textwidth]{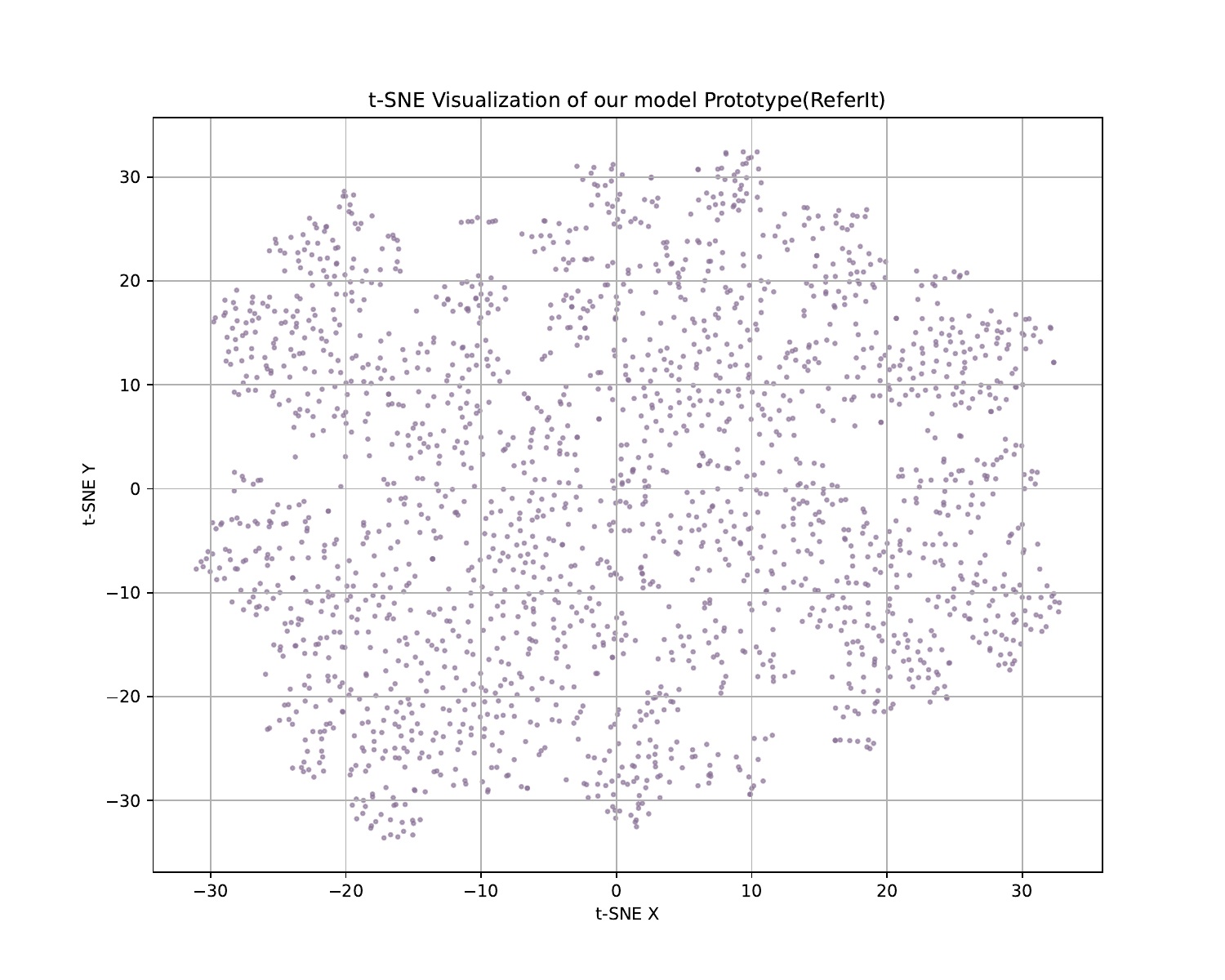}
		\caption{PAML: ReferIt}
		\label{fig:our_referit}
	\end{subfigure}
	\hfill
	\begin{subfigure}[b]{0.195\linewidth}
		\includegraphics[width=\textwidth]{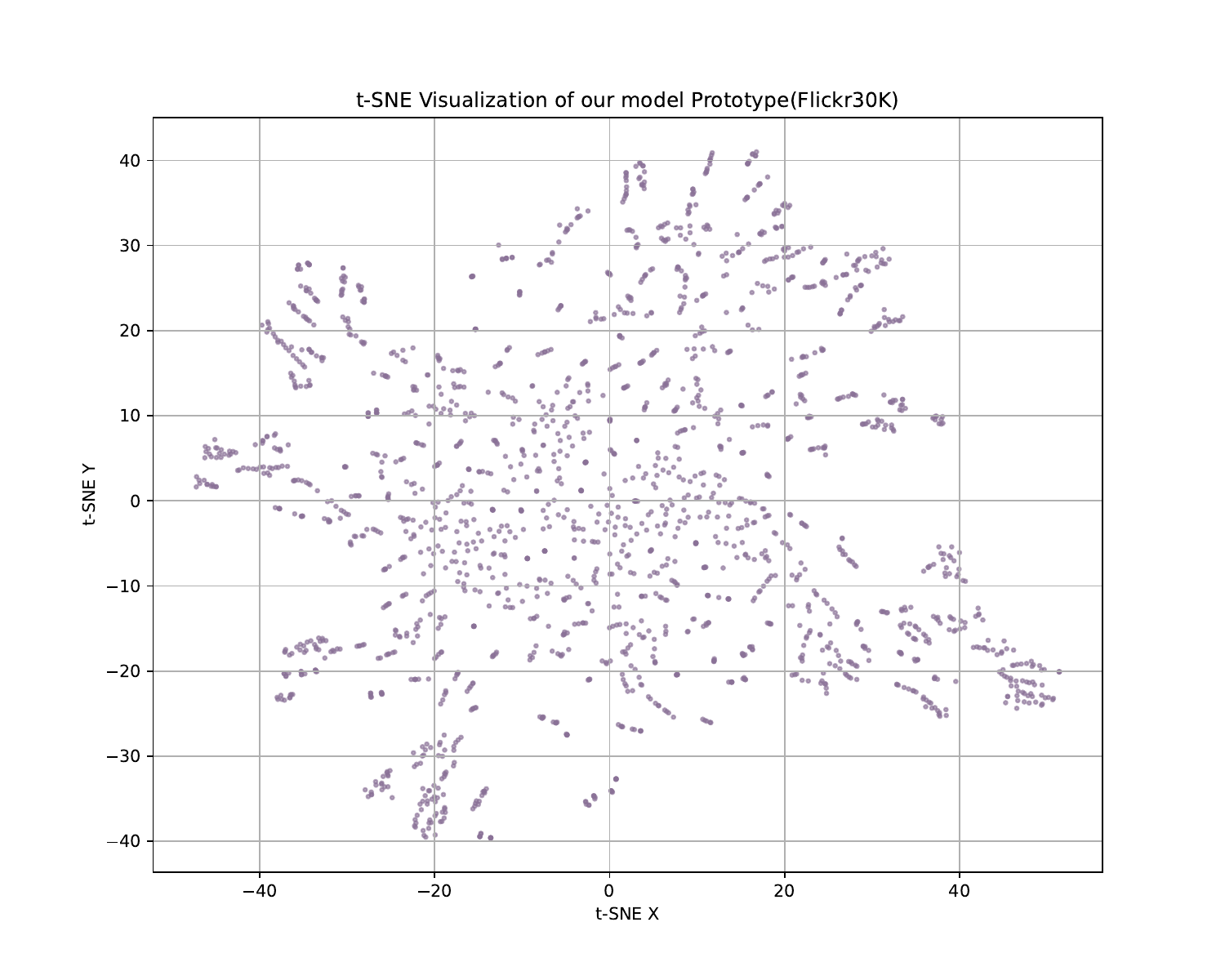}
		\caption{PAML: Flickr30k}
		\label{fig:our_flickr30k}
	\end{subfigure}
  
	\vspace{-0.1cm} 
  
	\begin{subfigure}[b]{0.195\linewidth}
		\includegraphics[width=\textwidth]{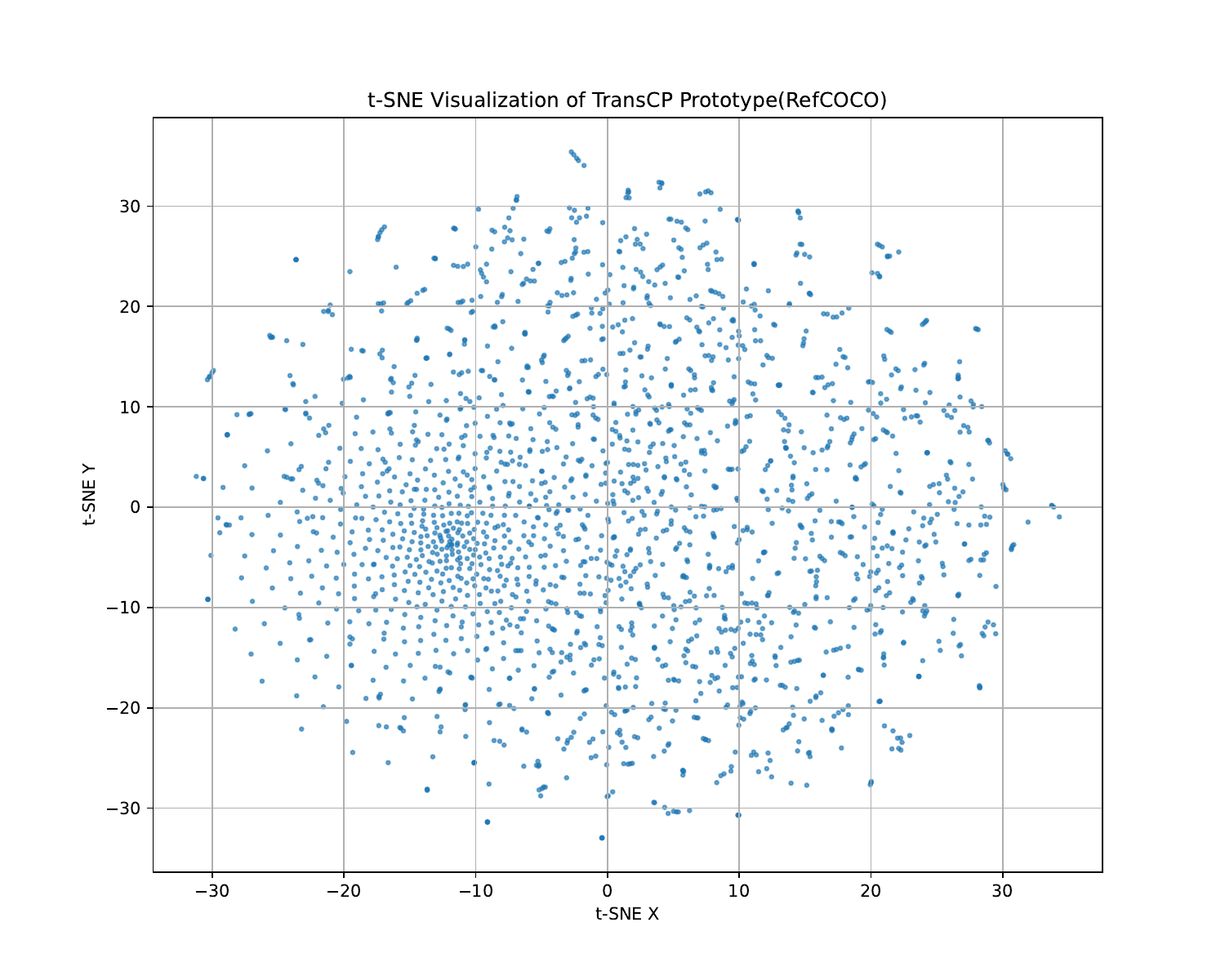}
		\caption{TransCP: RefCOCO}
		\label{fig:transcp_refcoco}
	\end{subfigure}
	\hfill
	\begin{subfigure}[b]{0.195\linewidth}
		\includegraphics[width=\textwidth]{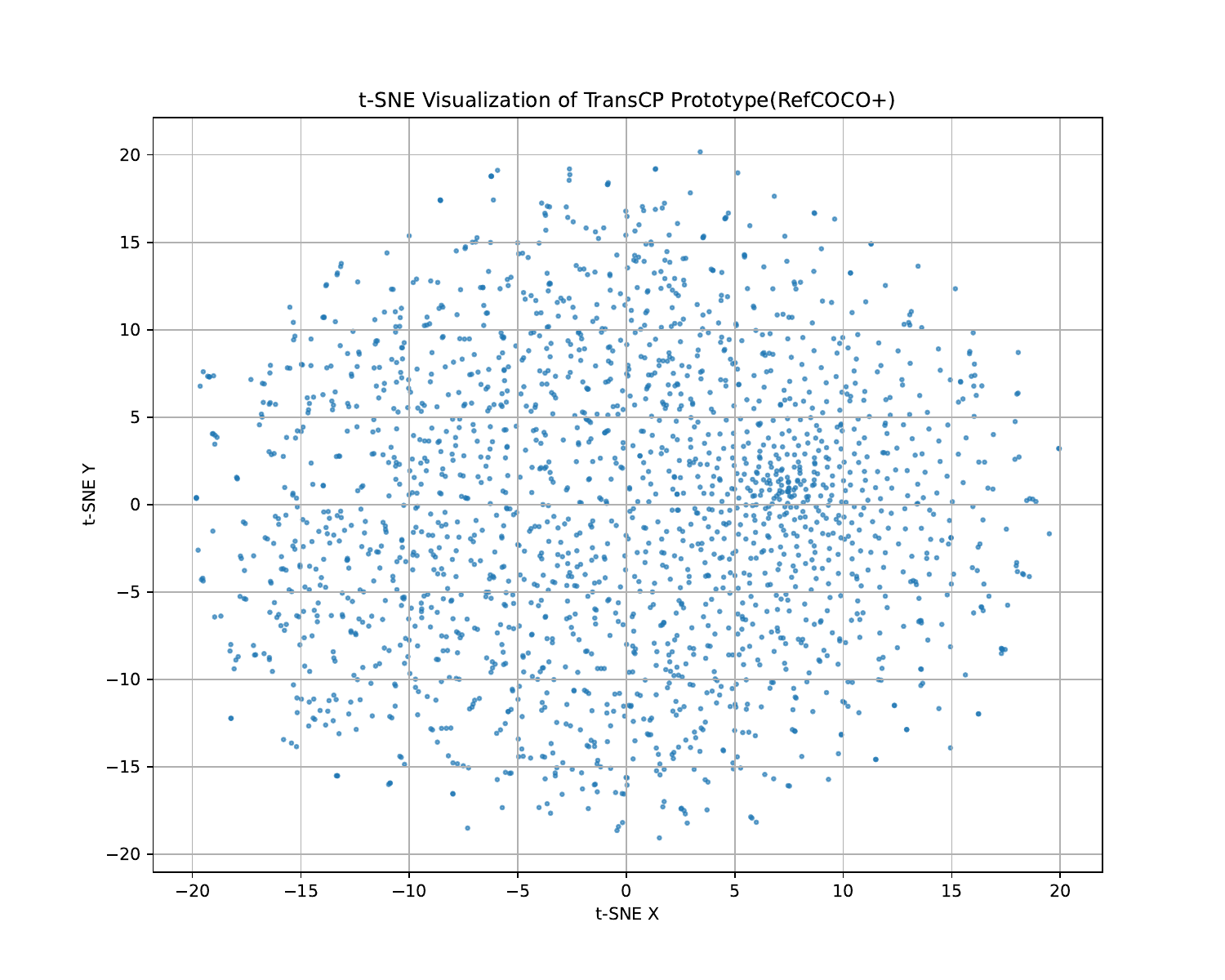}
		\caption{TransCP: RefCOCO+}
		\label{fig:transcp_refcocoplus}
	\end{subfigure}
	\hfill
	\begin{subfigure}[b]{0.195\linewidth}
		\includegraphics[width=\textwidth]{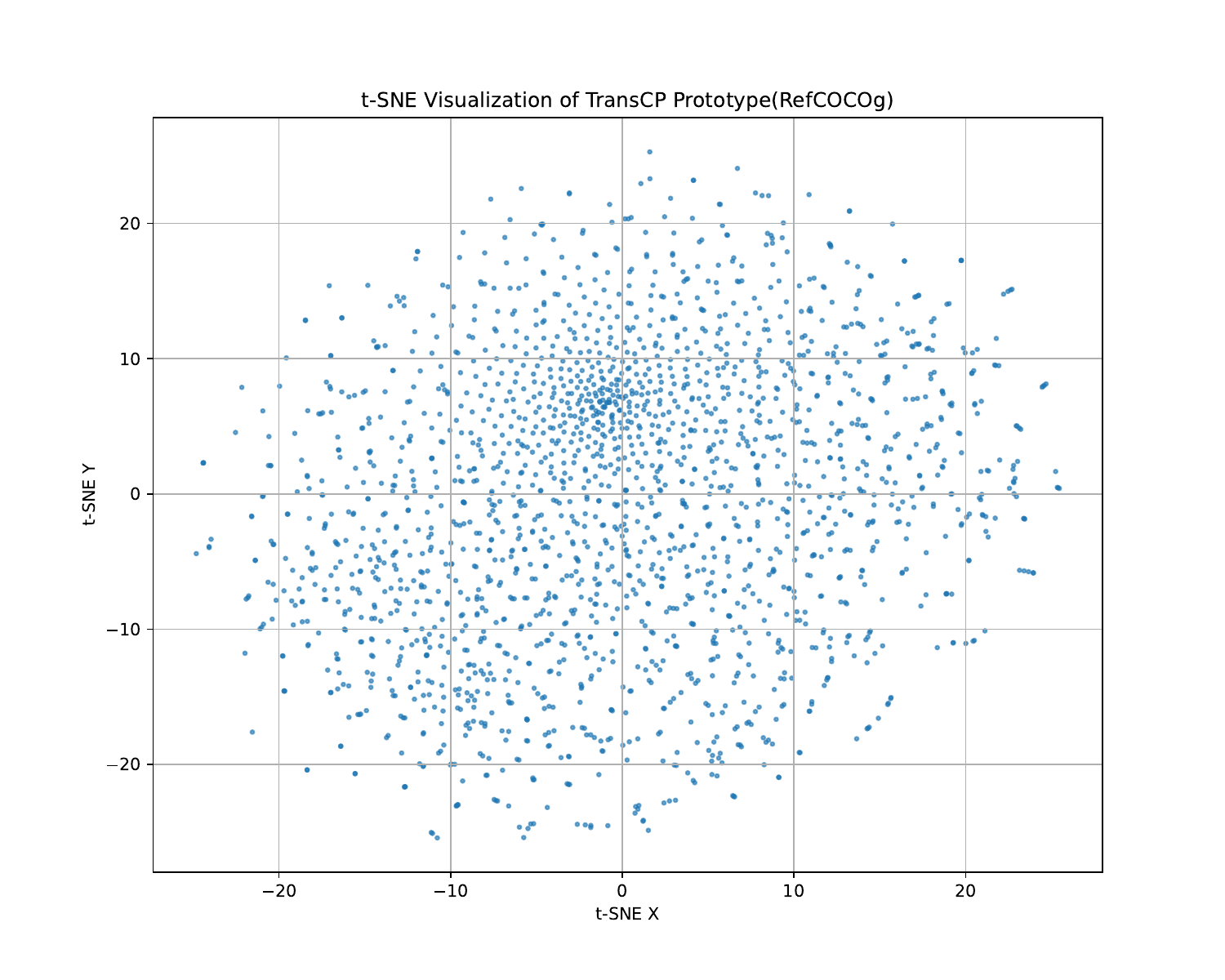}
		\caption{TransCP: RefCOCOg}
		\label{fig:transcp_refcocog}
	\end{subfigure}
	\hfill
	\begin{subfigure}[b]{0.195\linewidth}
		\includegraphics[width=\textwidth]{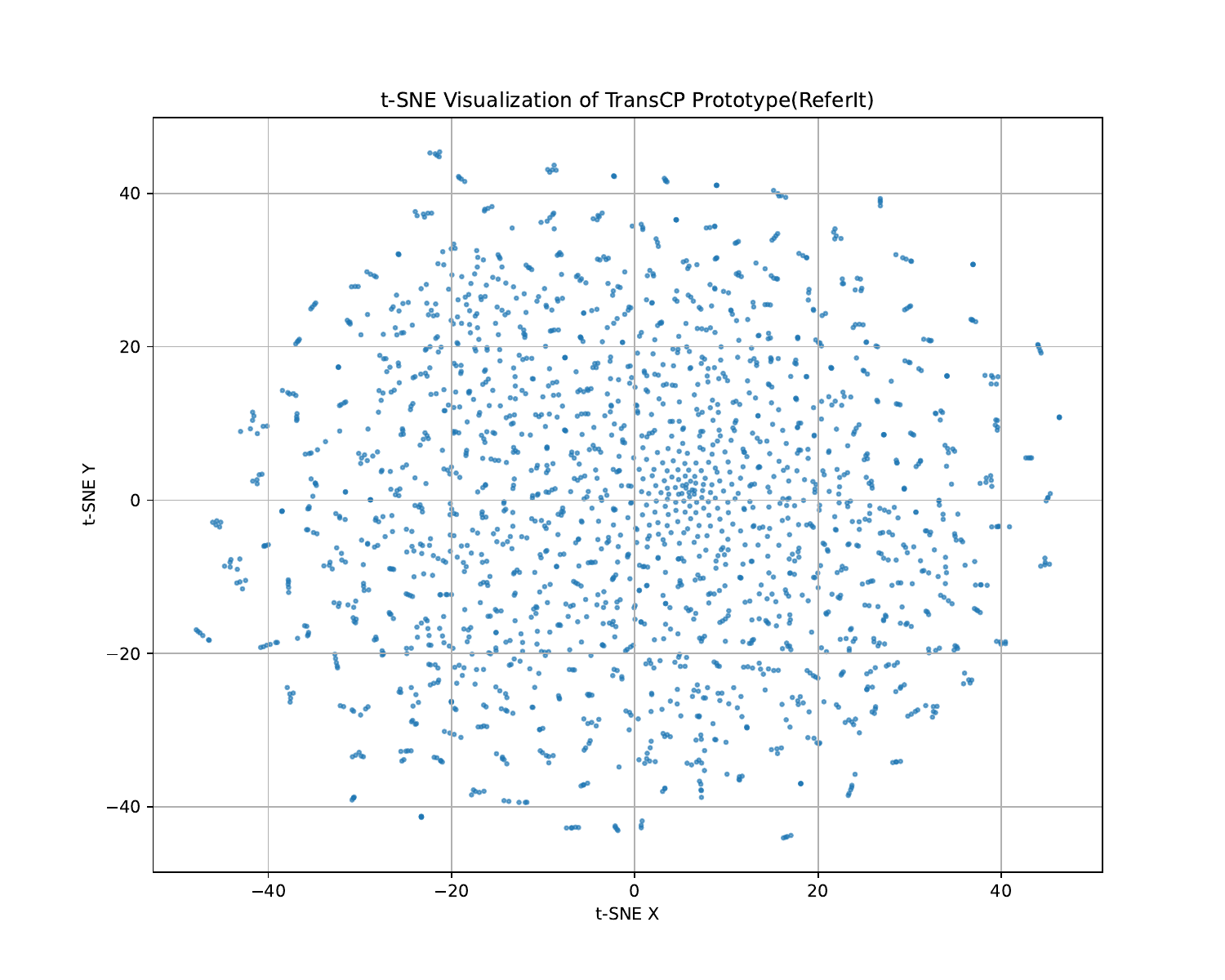}
		\caption{TransCP: ReferIt}
		\label{fig:transcp_referit}
	\end{subfigure}
	\hfill
	\begin{subfigure}[b]{0.195\linewidth}
		\includegraphics[width=\textwidth]{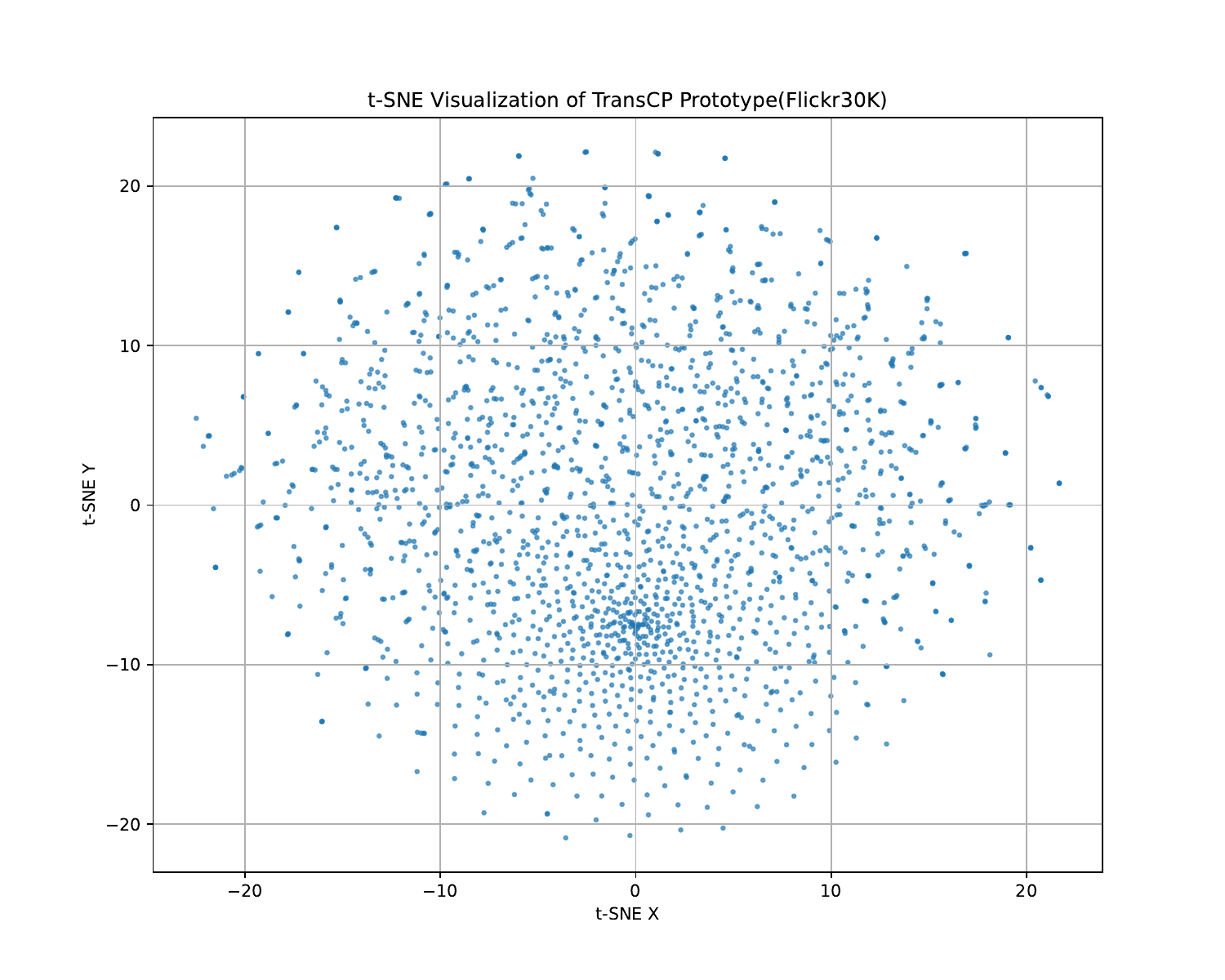}
		\caption{TransCP: Flickr30k}
		\label{fig:transcp_flickr30k}
	\end{subfigure}
  
	\caption{A t-SNE visualization is presented to compare
	the prototype bank distributions between our PAML framework (top row) and the TransCP baseline (bottom row) 
	across five benchmark datasets (left to right: RefCOCO, RefCOCO+, RefCOCOg, ReferIt, and Flickr30K).}
	\label{fig:comparison_grid}
  \end{figure*}

Eq.~\eqref{eq:gaussian}, Eq.~\eqref{eq:Laplacian} and Eq.~\eqref{eq:phivisual} describe the process where input $\phi_{sim}$ 
undergoes both Gaussian and Laplacian transformations, with the resulting outputs being combined via learnable weighting 
parameters $\lambda$ to produce $\phi_{v}$. The effectiveness of this operation is also verified through ablation experiments, 
with corresponding results shown in Table ~\ref{tab:ablation-transformation}.
When the learnable parameter $\lambda$ is disabled, we simply average inputs $\phi_{G}$ and $\phi_{L}$ with fixed 
weights of $0.5$. As shown in Table ~\ref{tab:ablation-transformation}, the model achieves its poorest performance 
($84.91\%$, $87.51\%$, and $82.64\%$ across different sets) when either transformation is omitted. The introduction of either 
Gaussian or Laplacian transformation yields consistent improvements - for instance, employing just the Gaussian transform 
boosts performance by $0.36\%$, $0.3\%$, and $0.2\%$ respectively. However, when both transformations are applied without the 
learnable parameter $\lambda$, the accuracy remains suboptimal, trailing the full configuration by $0.36\%$, $0.17\%$, and $0.18\%$.
These results substantiate two key findings: first, both transformations are essential for smoothing the data 
distribution $\phi_{sim}$ to reduce noise sensitivity and enhance robustness; second, the learnable parameter $\lambda$ 
is crucial for optimally balancing the complementary strengths of these transformations in the final output, As mentioned 
in Sec.~\ref{sec:Visual Discriminative Feature Encoder}, The Gaussian transformation emphasizes values near $1$, which 
correspond to strong similarities, by assigning them higher weights, thereby allowing the model to concentrate more on regions 
with high correlation between images and text. This helps improve the effectiveness of feature merging. Conversely, 
the Laplacian transformation captures the absolute differences in $\phi_{sim}$, making it more adept at managing outliers 
or anomalies. This results in increased stability, especially when the similarity scores are noisy or unreliable. 
Additionally, unlike the Gaussian approach, the Laplacian method distributes weights across both high and low similarity 
regions, preventing the model from disproportionately focusing only on highly similar areas.
Consequently, maintaining an equilibrium between the contributions of both transformations is crucial for model performance.

\begin{table}[t]
	\caption{Ablation study of different number of neighbor prototypes in standard scene}
	\small
	\begin{center}
		\scalebox{1}{
			\begin{tabular}{c | c c c}
				\hline
				\multicolumn{1}{c|}{Prototype} & \multicolumn{3}{c|}{RefCOCO}\\
				& val & testA & testB\\
				\hline
				\hline
				1 & 84.31 & 87.42 & 80.02 \\
				3 & 85.16 & 87.61 & 82.89 \\
				5 & 85.68 & 88.07 & 83.47 \\
				7 & 85.49 & 87.82 & 82.95 \\
				9 & 84.60 & 87.31 & 81.57 \\
				\hline
			\end{tabular}
		} 
	\end{center}
	\label{tab:ablation-prototype-number}
	\vspace{-0.10cm}
\end{table}

\begin{figure*}[t] 
   \centering
   \includegraphics[width=0.8\textwidth]{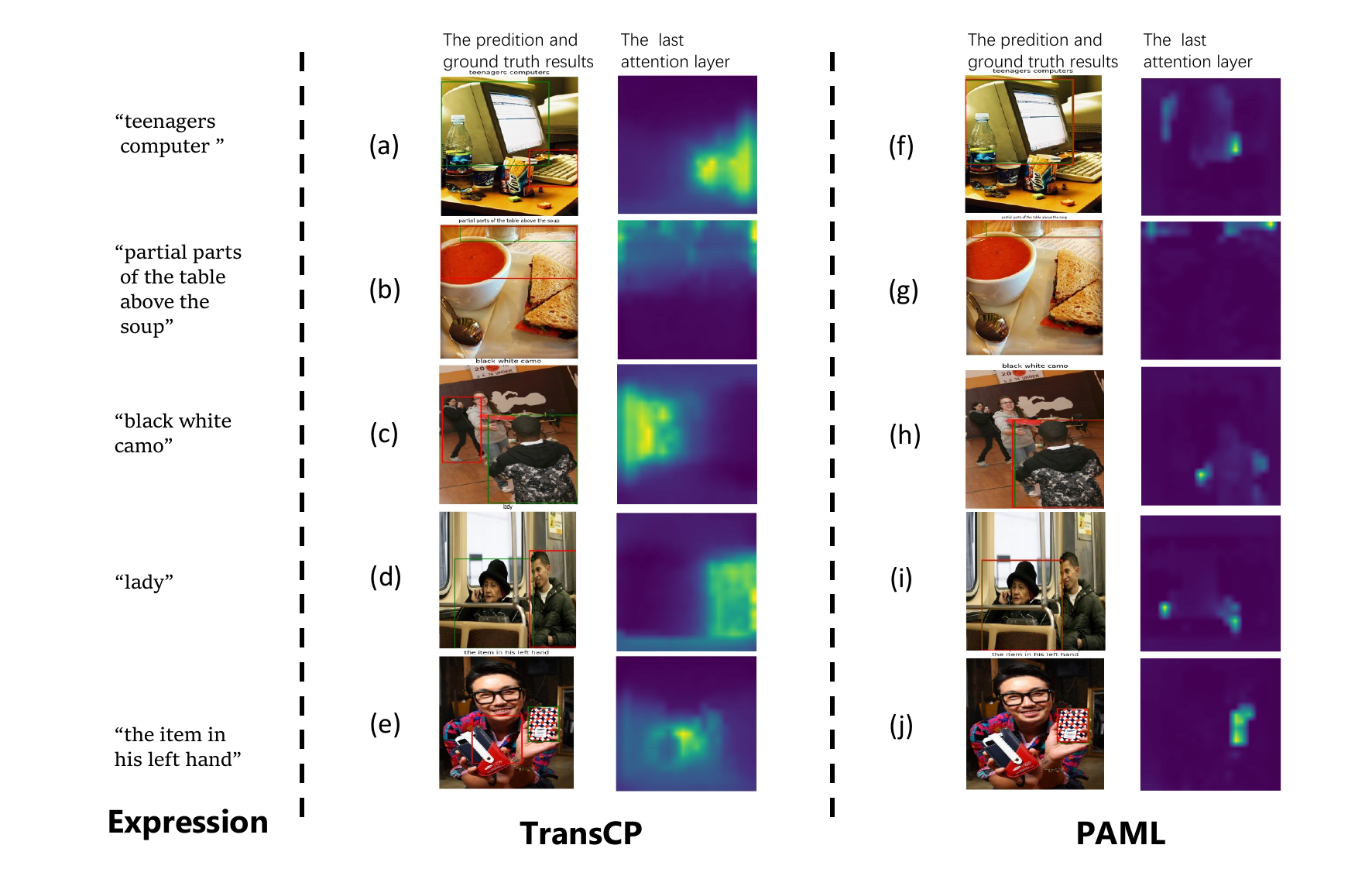} 
   \caption{presents a 
   comparative visualization of prediction results and final-layer attention maps between PAML (right column) 
   and TransCP (middle column), with corresponding expressions for each example(left column) in the standard 
   scene that train on RefCOCO training set and test on RefCOCO testA set. Red: prediction result; Green: ground truth.}
   \label{fig:standard-CAM}
 \end{figure*}

Our Multiple Neighbor Prototype Discovering and Inheriting module computes the input feature's corresponding prototype 
via a distance-based weighted summation of its 5 nearest prototypes. To analyze the effect of the number of prototypes 
involved in this operation, we conduct ablation experiments using $1, 3, 5, 7$, and $9$ prototypes. The model's performance on 
the RefCOCO benchmark under standard-scene settings is reported in Table ~\ref{tab:ablation-prototype-number}.
The experimental results demonstrate that incorporating the $5$ nearest prototypes achieves the best performance, while 
deviations—either reducing or increasing the number of prototypes—adversely affect model accuracy. For instance, when only $1$ 
nearest prototype is considered, the model exhibits performance declines of $1.37\%$, $0.65\%$, and $3.45\%$ on the val, 
testA, and testB splits, respectively, relative to the 5-prototype configuration. Conversely, expanding the prototype count to $9$ 
leads to reductions of $1.08\%$, $0.76\%$, and $1.90\%$ in accuracy across the same evaluation sets.
The experimental results demonstrate that relying solely on a single prototype may fail to comprehensively 
capture the diversity of input features, particularly when the input lies in the semantic boundary regions between 
multiple prototypes. Such single-prototype inheritance can lead to information loss.
The distance-weighted mechanism enables the model to dynamically adjust the contribution of each prototype, thereby 
achieving more precise fitting of the input feature distribution. For instance, local details may be predominantly 
captured by nearby prototypes, while the global structure is jointly characterized through the collaborative 
representation of multiple prototypes.
However, The inclusion of an excessive number of prototypes risks introducing irrelevant feature representations 
into the weighted aggregation, thereby compromising model accuracy. Consequently, determining an optimal number of 
nearest-neighbor prototypes is critical for maintaining performance as well.

To analyze the computational complexity of PAML, we conduct experiments 
to determine time cost during both training and inference phases.
The computational efficiency analysis measures training latency by averaging the execution time 
across $10$ stochastic forward-backward passes, while inference performance is evaluated solely 
through forward propagation timing. All comparative models (TransVG, TransCP, and PAML) were implemented 
on an NVIDIA GeForce RTX 4090 GPU with 24GB memory under identical experimental conditions (batch size = 1), 
using ReferIt test split as the evaluation benchmark, with quantitative results presented in Table~\ref{tab:time_cost}.
Our analysis reveals that although PAML exhibits moderately increased parameter counts ($+2\%$), training time ($+3.2\%$), 
and inference time ($+3.4\%$) compared to TransCP, these computational overheads remain within the same order of magnitude. 
Crucially, PAML delivers superior cost-effectiveness when considering its performance gains - achieving $5\%$ higher accuracy 
on the ReferIt test benchmark than TransCP. This evidence strongly suggests that the novel methods (e.g., multi-prototype 
mechanism, multi-stage fusion) introduce only marginal computational complexity while significantly enhancing model capability.

\begin{table}[t!]
	\centering
	\caption{Comparison of the time cost and parameters with the baselines on the test split of ReferIt dataset.}
	\label{tab:time_cost}
	\begin{tabular}{lccc}
	\hline
	\textbf{Models} & \textbf{Params. (M)} & \textbf{Training (ms)} & \textbf{Inference (ms)} \\
	\hline
	\hline
	    TransVG & 150.96 & 122.79 & 37.22 \\
		TransCP & 160.43 & 119.59 & 39.89 \\
		PAML    & 163.57 & 123.37 & 41.25 \\

	\hline
	\end{tabular}
\end{table}

\subsection{Visualization}
\label{sec:Visualization}
\subsubsection{Prototype Bank}
\label{sec:Prototype Bank}

\begin{figure*}[htbp] 
    \centering
    \includegraphics[width=0.8\textwidth]{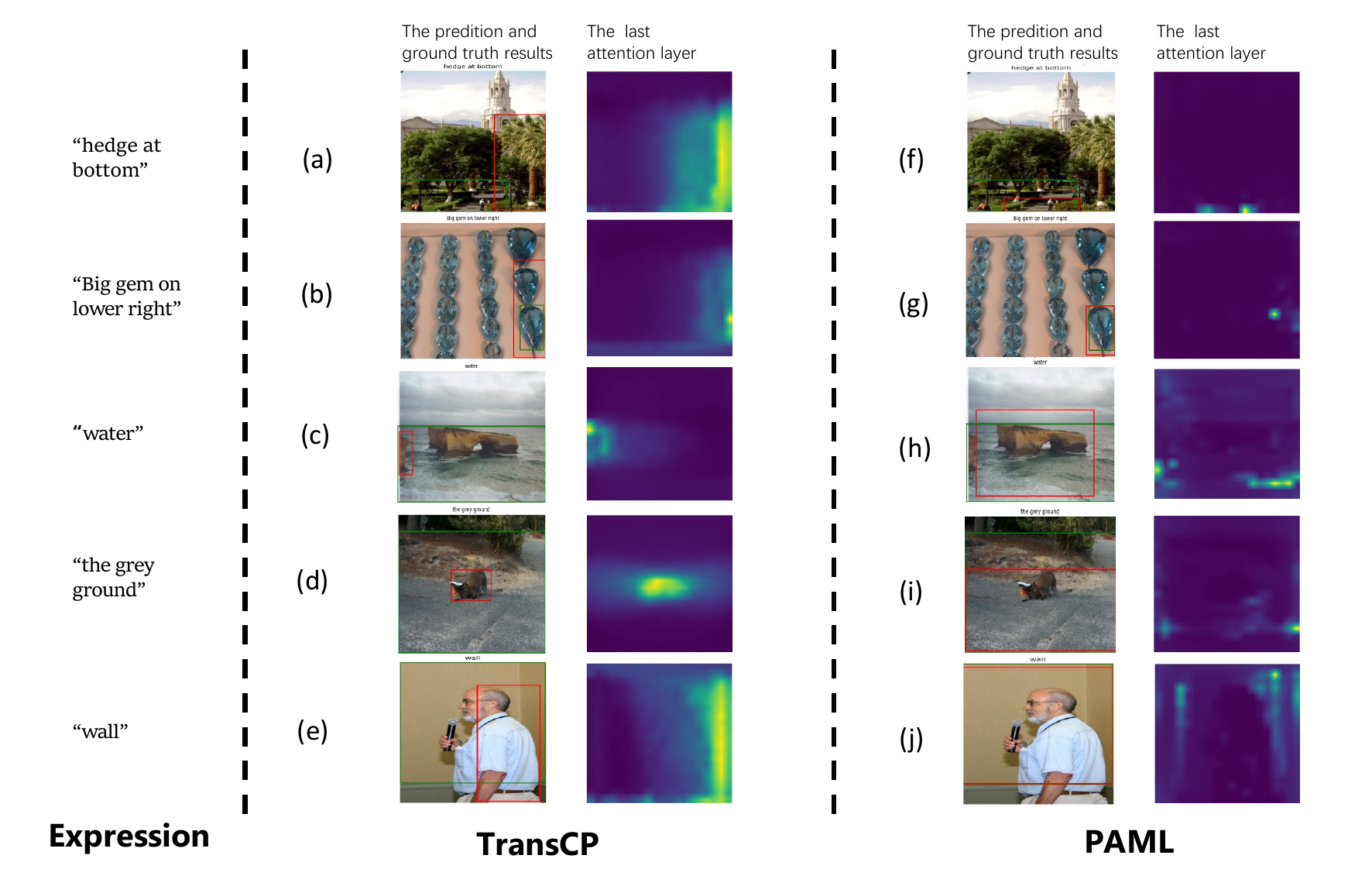} 
    \caption{presents a comparative visualization of prediction results and final-layer attention maps between PAML (right column) and TransCP (middle column), with corresponding
expressions for each example(left column) in the open-vocabulary scene that train on RefCOCO training set and test on ReferIt val set. Red: prediction result;
Green: ground truth.}
    \label{fig:open-CAM}
\end{figure*}

For a more comprehensive understanding of the prototype bank's underlying mechanism, we 
visualize the prototype banks of our model and the TransCP model, both trained on five distinct datasets, 
by applying t-SNE to reduce their dimensionality from $2048\times768$ to $2048\times2$. The resulting 2D projections 
are shown in Fig. \ref{fig:comparison_grid}.
As shown in the Fig. \ref{fig:comparison_grid}, although we predefined the prototype bank size as $2048$, 
the actual number of points projected onto the 2D plane is significantly fewer than $2048$. This indicates 
that some prototype features gradually converge in the semantic space during training and can ultimately be 
considered as identical prototypes. This aligns with our expectations, since the actual number of distinct 
prototypes within the dataset is necessarily smaller than the predefined bank size of 2048. The number of prototypes 
is closely related to the training dataset characteristics. For instance, the ReferIt dataset typically contains images 
with numerous targets and descriptions involving complex spatial relationships, whereas Flickr30K primarily consists of 
humans and common object categories. Consequently, ReferIt theoretically requires more prototypes. 
Fig. \ref{fig:our_refcocog} and Fig. \ref{fig:our_flickr30k} demonstrate that the prototype bank trained on ReferIt exhibits 
a more dispersed distribution compared to that trained on Flickr30K, representing a greater number of distinct prototypes, 
which aligns with our theoretical analysis.
Furthermore, the prototype distributions demonstrate higher density in our approach relative to TransCP, suggesting more 
effective prototype representation learning. This characteristic fundamentally accounts for our method's performance advantages 
in open-vocabulary scene.

\subsubsection{Attention map}
\label{Attention map}

To further highlight the strengths of our proposed model, we generate attention map visualizations of the 
attention scores corresponding to the [REG] tokens in the final attention layer, comparing our approach with the TransCP model.
We select representative examples from both the standard-scene and open-vocabulary scene settings, which are subsequently 
visualized in Fig. \ref{fig:standard-CAM} and Fig. \ref{fig:open-CAM}, respectively. All images and referring expressions in 
Fig. \ref{fig:standard-CAM} are sampled from the RefCOCO dataset (val, testA, and testB splits), while the model was 
exclusively trained on the RefCOCO train split. 
For Fig. \ref{fig:open-CAM}, the visual-grounding pairs originate from the ReferIt dataset (val and test splits), 
though the model maintains the same training protocol using only RefCOCO training data.

As demonstrated in 
Fig. \ref{fig:standard-CAM}(a), (f), (d), (i), and Fig. \ref{fig:open-CAM}(c), (h), (e), (j), 
our model exhibits superior fine-grained word comprehension compared to TransCP. For instance, it accurately 
localizes regions corresponding to specific words such as "lady" (identifying females in images) and "water" (detecting aqueous regions).

Furthermore, Fig. \ref{fig:standard-CAM}(b), (g), (e), (j) and Fig. \ref{fig:open-CAM}(a), (f), (b), (g) highlight 
our model's enhanced spatial reasoning capabilities. Notably, the localization task in Fig. 3(e), (j) is particularly 
challenging, as it requires interpreting object positions from the human perspective (e.g., held items) rather than the image's 
global viewpoint. Our model’s precise predictions in these cases underscore its robust understanding of linguistically 
specified spatial relationships.

Additionally, Fig. \ref{fig:standard-CAM}(c), (h) and Fig. \ref{fig:open-CAM}(d), (i) reveal our model's improved semantic 
accuracy for domain-specific vocabulary. For example, given the phrase "black white camo", TransCP erroneously associates 
it with a "female in black clothing holding a white phone", failing to capture the true meaning (camouflage pattern). 
In contrast, our model correctly interprets the term and localizes the target object.

Meanwhile, Regarding the attention maps, we observe that, compared to TransCP, our model generates attention maps that 
are more focused on specific regions and already exhibit approximate contours of the target objects, achieving a 
segmentation-like effect. This indicates a unique advantage of our model in the task of target localization.

\subsubsection{Irrelevant context suppression}
\label{Irrelevant context suppression}
\begin{figure*}[htbp] 
    \centering
    \includegraphics[width=\textwidth]{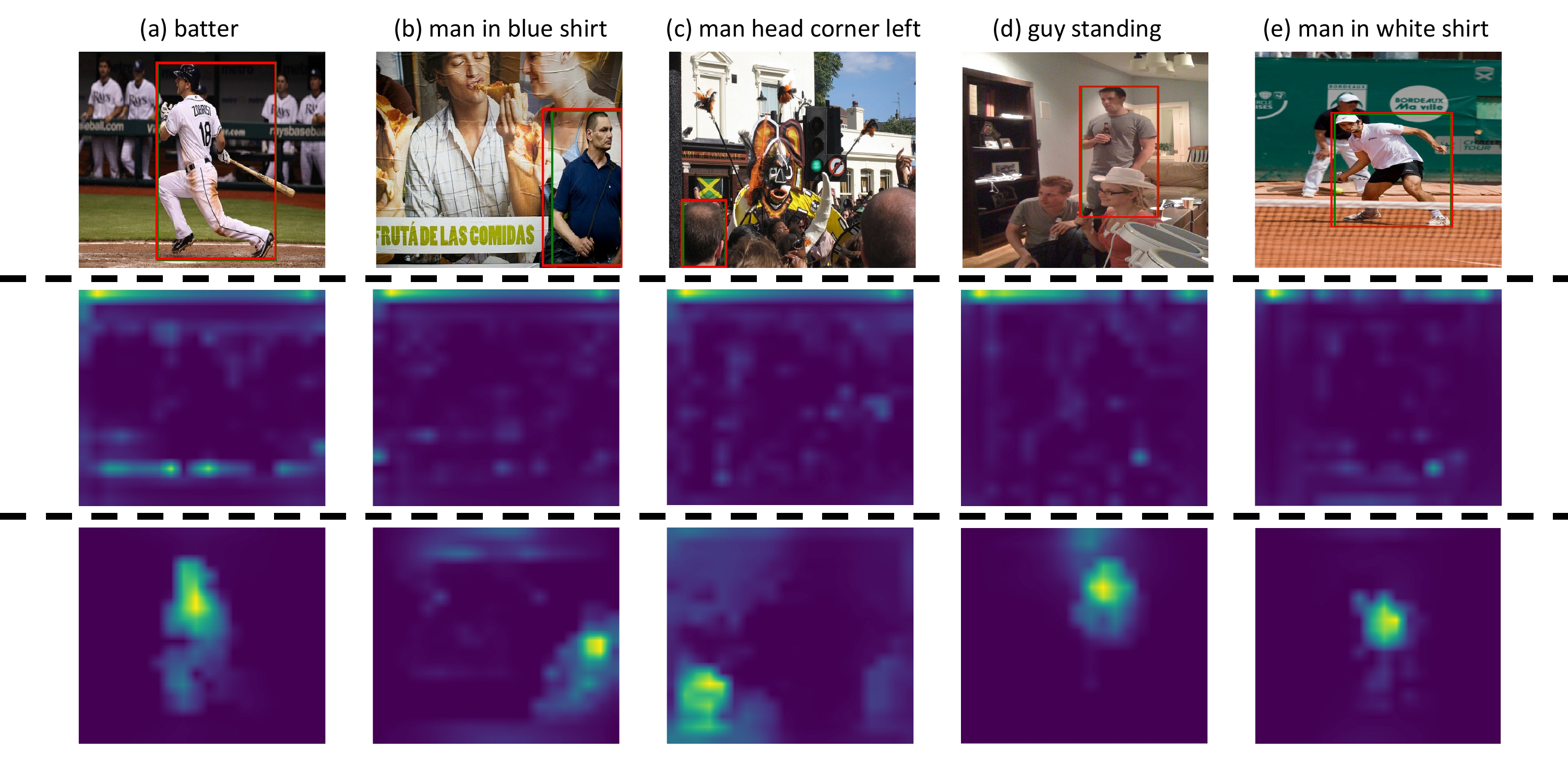} 
    \caption{presents a comparative visualization of attention maps between the 
	final attention layer before the Visual Discriminative Feature Encoder (middle row), and 
	the final attention layer of the Visual Discriminative Feature Encoder (bottom row). The corresponding 
	input image with prediction results and textual expressions are displayed in the top row. Red: prediction 
	result; Green: ground truth.}
    \label{fig:context_suppress-CAM}
\end{figure*}
In Sec.~\ref{sec:Visual Discriminative Feature Encoder}, we introduced that the Visual Discriminative Feature 
Encoder enhances salient objects while suppressing irrelevant contextual information. To verify this functionality, 
we extract attention weights from both the final attention layer preceding this module (i.e., the last 
layer of ALBEF's Visual Encoder) and the final attention layer within the Visual Discriminative Feature Encoder, 
visualizing them as attention maps in 
Fig.~\ref{fig:context_suppress-CAM} for comparative analysis.
From the attention maps, we can observe that before applying the Visual Discriminative Feature Encoder, 
the attention weights are diffusely distributed, whereas after processing through the Visual Discriminative Feature Encoder, 
the model successfully highlights language-relevant objects in the image while significantly suppressing irrelevant environmental 
elements for the localization task.
For instance, in Fig.~\ref{fig:context_suppress-CAM}(d),(e), the Visual Discriminative Feature Encoder effectively highlights 
the approximate locations of target objects (i.e. "guy", "man") while suppressing irrelevant background 
information (e.g. "bookshelf", "billboard"), establishing a robust foundation for subsequent modules to further 
refine both the size and contours of these objects.

\section{Conclusion}
We propose PAML, a novel framework for open-vocabulary visual grounding that enhances cross-modal alignment, 
prototype-based reasoning, and deep multimodal fusion. By integrating ALBEF for robust feature encoding, a Visual 
Discriminative Feature Encoder for salient object enhancement, Multiple Neighbor Prototype Discovering and Inheriting 
for generalizable prototype learning and retrieval, and a Multi-Stage Decoder for thorough cross-modal integration, 
PAML achieves state-of-the-art performance in both standard and open-vocabulary scenes. Extensive experiments on five 
benchmarks validate its superiority, particularly in handling unseen objects and complex queries.

Future improvements could focus on several aspects to enhance model performance. The current 
framework employs ALBEF for data encoding, which could be replaced by more advanced vision-language models as they 
emerge to achieve better performance. Additionally, the prototype bank uses a fixed number of predefined prototypes, 
whereas adopting a dynamic mechanism to adjust the prototype count based on training data characteristics may yield 
more adaptive representations. Furthermore, the Transformer-based approach relies solely on REG tokens for bounding box 
regression, raising questions about whether this design enables sufficient fusion with multimodal features. Future work 
could explore alternative regression strategies to improve localization accuracy while maintaining effective multimodal integration.

\section*{CRediT authorship contribution statement}
\textbf{Jiangnan Xie}: Methodology, Validation, Data curation, Formal analysis, Investigation, Writing - original draft.
\textbf{Xiaolong Zheng}: Methodology, Investigation, Formal analysis, Supervision, Visualization, Writing - original draft, Writing - review \& editing. 
\textbf{Liang Zheng}: Investigation, Formal analysis.

\section*{Declaration of competing interest}
The authors declare that they have no known competing financial interests or personal relationships that could have appeared to influence the work reported in this paper.

\section*{Data availability}
The datasets employed in this study were obtained from publicly available repositories, accessible via their corresponding references. The code used in this study is available at https://github.com/plankXie/PAML.

\section*{Acknowledgements}
This work was supported by the National Key R\&D Program of China (Grant No. 2024YFB4207200).

{\small
	\bibliographystyle{ieee_fullname}
	\bibliography{egbib}

\begin{thebibliography}{10}\itemsep=-1pt

\bibitem{anderson2018vision}
Peter Anderson, Qi Wu, Damien Teney, Jake Bruce, Mark Johnson, Niko S{\"u}nderhauf, Ian Reid, Stephen Gould, and Anton Van Den~Hengel.
\newblock Vision-and-language navigation: Interpreting visually-grounded navigation instructions in real environments.
\newblock In {\em Proceedings of the IEEE conference on computer vision and pattern recognition}, pages 3674--3683, 2018.

\bibitem{antol2015vqa}
Stanislaw Antol, Aishwarya Agrawal, Jiasen Lu, Margaret Mitchell, Dhruv Batra, C Lawrence~Zitnick, and Devi Parikh.
\newblock Vqa: Visual question answering.
\newblock In {\em ICCV}, pages 2425--2433, 2015.

\bibitem{bao2022vlmo}
Hangbo Bao, Wenhui Wang, Li Dong, Qiang Liu, Owais~Khan Mohammed, Kriti Aggarwal, Subhojit Som, Songhao Piao, and Furu Wei.
\newblock Vlmo: Unified vision-language pre-training with mixture-of-modality-experts.
\newblock {\em Advances in Neural Information Processing Systems}, 35:32897--32912, 2022.

\bibitem{belkin2004semi}
Mikhail Belkin and Partha Niyogi.
\newblock Semi-supervised learning on riemannian manifolds.
\newblock {\em Machine learning}, 56:209--239, 2004.

\bibitem{chen2018knowledge}
Kan Chen, Jiyang Gao, and Ram Nevatia.
\newblock Knowledge aided consistency for weakly supervised phrase grounding.
\newblock In {\em Proceedings of the IEEE conference on computer vision and pattern recognition}, pages 4042--4050, 2018.

\bibitem{chen2021ref}
Long Chen, Wenbo Ma, Jun Xiao, Hanwang Zhang, and Shih-Fu Chang.
\newblock Ref-nms: Breaking proposal bottlenecks in two-stage referring expression grounding.
\newblock In {\em Proceedings of the AAAI conference on artificial intelligence}, volume~35, pages 1036--1044, 2021.

\bibitem{chen2020counterfactual}
Long Chen, Xin Yan, Jun Xiao, Hanwang Zhang, Shiliang Pu, and Yueting Zhuang.
\newblock Counterfactual samples synthesizing for robust visual question answering.
\newblock In {\em CVPR}, pages 10800--10809, 2020.

\bibitem{chen2018real}
Xinpeng Chen, Lin Ma, Jingyuan Chen, Zequn Jie, Wei Liu, and Jiebo Luo.
\newblock Real-time referring expression comprehension by single-stage grounding network.
\newblock {\em arXiv preprint arXiv:1812.03426}, 2018.

\bibitem{chen2021zero}
Zhuo Chen, Jiaoyan Chen, Yuxia Geng, Jeff~Z Pan, Zonggang Yuan, and Huajun Chen.
\newblock Zero-shot visual question answering using knowledge graph.
\newblock In {\em The Semantic Web--ISWC 2021: 20th International Semantic Web Conference, ISWC 2021, Virtual Event, October 24--28, 2021, Proceedings 20}, pages 146--162. Springer, 2021.

\bibitem{Dai_Li_Zhuang_Zhang_Yang_2025}
Ming Dai, Jian Li, Jiedong Zhuang, Xian Zhang, and Wankou Yang.
\newblock Multi-task visual grounding with coarse-to-fine consistency constraints.
\newblock {\em Proceedings of the AAAI Conference on Artificial Intelligence}, 39(3):2618--2626, Apr. 2025.

\bibitem{dai2024simvg}
Ming Dai, Lingfeng Yang, Yihao Xu, Zhenhua Feng, and Wankou Yang.
\newblock Simvg: A simple framework for visual grounding with decoupled multi-modal fusion.
\newblock {\em Advances in neural information processing systems}, 37:121670--121698, 2024.

\bibitem{datta2019align2ground}
Samyak Datta, Karan Sikka, Anirban Roy, Karuna Ahuja, Devi Parikh, and Ajay Divakaran.
\newblock Align2ground: Weakly supervised phrase grounding guided by image-caption alignment.
\newblock In {\em Proceedings of the IEEE/CVF international conference on computer vision}, pages 2601--2610, 2019.

\bibitem{dempster1977maximum}
Arthur~P Dempster, Nan~M Laird, and Donald~B Rubin.
\newblock Maximum likelihood from incomplete data via the em algorithm.
\newblock {\em Journal of the royal statistical society: series B (methodological)}, 39(1):1--22, 1977.

\bibitem{deng2018visual}
Chaorui Deng, Qi Wu, Qingyao Wu, Fuyuan Hu, Fan Lyu, and Mingkui Tan.
\newblock Visual grounding via accumulated attention.
\newblock In {\em Proceedings of the IEEE conference on computer vision and pattern recognition}, pages 7746--7755, 2018.

\bibitem{deng2021transvg}
Jiajun Deng, Zhengyuan Yang, Tianlang Chen, Wengang Zhou, and Houqiang Li.
\newblock Transvg: End-to-end visual grounding with transformers.
\newblock In {\em Proceedings of the IEEE/CVF International Conference on Computer Vision}, pages 1769--1779, 2021.

\bibitem{deng2023transvg++}
Jiajun Deng, Zhengyuan Yang, Daqing Liu, Tianlang Chen, Wengang Zhou, Yanyong Zhang, Houqiang Li, and Wanli Ouyang.
\newblock Transvg++: End-to-end visual grounding with language conditioned vision transformer.
\newblock {\em IEEE transactions on pattern analysis and machine intelligence}, 45(11):13636--13652, 2023.

\bibitem{driess2023palm}
Danny Driess, Fei Xia, Mehdi~SM Sajjadi, Corey Lynch, Aakanksha Chowdhery, Ayzaan Wahid, Jonathan Tompson, Quan Vuong, Tianhe Yu, Wenlong Huang, et~al.
\newblock Palm-e: An embodied multimodal language model.
\newblock 2023.

\bibitem{du2022visual}
Ye Du, Zehua Fu, Qingjie Liu, and Yunhong Wang.
\newblock Visual grounding with transformers.
\newblock In {\em Proceedings of the International Conference on Multimedia and Expo}, 2022.

\bibitem{escalante2010segmented}
Hugo~Jair Escalante, Carlos~A Hern{\'a}ndez, Jesus~A Gonzalez, Aurelio L{\'o}pez-L{\'o}pez, Manuel Montes, Eduardo~F Morales, L~Enrique Sucar, Luis Villase{\~n}or, and Michael Grubinger.
\newblock The segmented and annotated iapr tc-12 benchmark.
\newblock {\em CVIU}, 114:419--428, 2010.

\bibitem{gu2022openvocabulary}
Xiuye Gu, Tsung-Yi Lin, Weicheng Kuo, and Yin Cui.
\newblock Open-vocabulary object detection via vision and language knowledge distillation.
\newblock In {\em International Conference on Learning Representations}, 2022.

\bibitem{he2024improved}
Ruozhen He, Paola Cascante-Bonilla, Ziyan Yang, Alexander~C Berg, and Vicente Ordonez.
\newblock Improved visual grounding through self-consistent explanations.
\newblock In {\em Proceedings of the IEEE/CVF Conference on Computer Vision and Pattern Recognition}, pages 13095--13105, 2024.

\bibitem{hong2019learning}
Richang Hong, Daqing Liu, Xiaoyu Mo, Xiangnan He, and Hanwang Zhang.
\newblock Learning to compose and reason with language tree structures for visual grounding.
\newblock {\em TPAMI}, 2019.

\bibitem{hu2017modeling}
Ronghang Hu, Marcus Rohrbach, Jacob Andreas, Trevor Darrell, and Kate Saenko.
\newblock Modeling relationships in referential expressions with compositional modular networks.
\newblock In {\em CVPR}, pages 1115--1124, 2017.

\bibitem{hu2016natural}
Ronghang Hu, Huazhe Xu, Marcus Rohrbach, Jiashi Feng, Kate Saenko, and Trevor Darrell.
\newblock Natural language object retrieval.
\newblock In {\em CVPR}, pages 4555--4564, 2016.

\bibitem{hu2021vivo}
Xiaowei Hu, Xi Yin, Kevin Lin, Lei Zhang, Jianfeng Gao, Lijuan Wang, and Zicheng Liu.
\newblock Vivo: Visual vocabulary pre-training for novel object captioning.
\newblock In {\em proceedings of the AAAI conference on artificial intelligence}, volume~35, pages 1575--1583, 2021.

\bibitem{huang2021look}
Binbin Huang, Dongze Lian, Weixin Luo, and Shenghua Gao.
\newblock Look before you leap: Learning landmark features for one-stage visual grounding.
\newblock In {\em Proceedings of the IEEE/CVF conference on computer vision and pattern recognition}, pages 16888--16897, 2021.

\bibitem{huang2019multi}
Pingping Huang, Jianhui Huang, Yuqing Guo, Min Qiao, and Yong Zhu.
\newblock Multi-grained attention with object-level grounding for visual question answering.
\newblock In {\em Proceedings of the 57th Annual Meeting of the Association for Computational Linguistics}, pages 3595--3600, 2019.

\bibitem{jiang2022pseudo}
Haojun Jiang, Yuanze Lin, Dongchen Han, Shiji Song, and Gao Huang.
\newblock Pseudo-q: Generating pseudo language queries for visual grounding.
\newblock In {\em Proceedings of the IEEE/CVF Conference on Computer Vision and Pattern Recognition}, pages 15513--15523, 2022.

\bibitem{jin2023pseudo}
Jianglin Jin, Jiabo Ye, Xin Lin, and Liang He.
\newblock Pseudo-query generation for semi-supervised visual grounding with knowledge distillation.
\newblock In {\em ICASSP 2023-2023 IEEE International Conference on Acoustics, Speech and Signal Processing (ICASSP)}, pages 1--5. IEEE, 2023.

\bibitem{kazemzadeh2014referitgame}
Sahar Kazemzadeh, Vicente Ordonez, Mark Matten, and Tamara Berg.
\newblock Referitgame: Referring to objects in photographs of natural scenes.
\newblock In {\em EMNLP}, 2014.

\bibitem{kovvuri2018pirc}
Rama Kovvuri and Ram Nevatia.
\newblock Pirc net: Using proposal indexing, relationships and context for phrase grounding.
\newblock In {\em ACCV}, pages 451--467, 2018.

\bibitem{li2020transferrable}
Aoxue Li, Zhiwu Lu, Jiechao Guan, Tao Xiang, Liwei Wang, and Ji-Rong Wen.
\newblock Transferrable feature and projection learning with class hierarchy for zero-shot learning.
\newblock {\em International Journal of Computer Vision}, 128:2810--2827, 2020.

\bibitem{li2022blip}
Junnan Li, Dongxu Li, Caiming Xiong, and Steven Hoi.
\newblock Blip: Bootstrapping language-image pre-training for unified vision-language understanding and generation.
\newblock In {\em International conference on machine learning}, pages 12888--12900. PMLR, 2022.

\bibitem{li2021align}
Junnan Li, Ramprasaath Selvaraju, Akhilesh Gotmare, Shafiq Joty, Caiming Xiong, and Steven Chu~Hong Hoi.
\newblock Align before fuse: Vision and language representation learning with momentum distillation.
\newblock {\em Advances in neural information processing systems}, 34:9694--9705, 2021.

\bibitem{li2021referring}
Muchen Li and Leonid Sigal.
\newblock Referring transformer: A one-step approach to multi-task visual grounding.
\newblock {\em Advances in neural information processing systems}, 34:19652--19664, 2021.

\bibitem{li2020oscar}
Xiujun Li, Xi Yin, Chunyuan Li, Pengchuan Zhang, Xiaowei Hu, Lei Zhang, Lijuan Wang, Houdong Hu, Li Dong, Furu Wei, et~al.
\newblock Oscar: Object-semantics aligned pre-training for vision-language tasks.
\newblock In {\em ECCV}, pages 121--137, 2020.

\bibitem{li2018visual}
Yikang Li, Nan Duan, Bolei Zhou, Xiao Chu, Wanli Ouyang, Xiaogang Wang, and Ming Zhou.
\newblock Visual question generation as dual task of visual question answering.
\newblock In {\em Proceedings of the IEEE conference on computer vision and pattern recognition}, pages 6116--6124, 2018.

\bibitem{liao2020real}
Yue Liao, Si Liu, Guanbin Li, Fei Wang, Yanjie Chen, Chen Qian, and Bo Li.
\newblock A real-time cross-modality correlation filtering method for referring expression comprehension.
\newblock In {\em CVPR}, pages 10880--10889, 2020.

\bibitem{liao2022progressive}
Yue Liao, Aixi Zhang, Zhiyuan Chen, Tianrui Hui, and Si Liu.
\newblock Progressive language-customized visual feature learning for one-stage visual grounding.
\newblock {\em IEEE Transactions on Image Processing}, 31:4266--4277, 2022.

\bibitem{lin2014microsoft}
Tsung-Yi Lin, Michael Maire, Serge Belongie, James Hays, Pietro Perona, Deva Ramanan, Piotr Doll{\'a}r, and C~Lawrence Zitnick.
\newblock Microsoft coco: Common objects in context.
\newblock In {\em ECCV}, pages 740--755, 2014.

\bibitem{liu2019learning}
Daqing Liu, Hanwang Zhang, Feng Wu, and Zheng-Jun Zha.
\newblock Learning to assemble neural module tree networks for visual grounding.
\newblock In {\em ICCV}, pages 4673--4682, 2019.

\bibitem{mao2016generation}
Junhua Mao, Jonathan Huang, Alexander Toshev, Oana Camburu, Alan~L Yuille, and Kevin Murphy.
\newblock Generation and comprehension of unambiguous object descriptions.
\newblock In {\em CVPR}, pages 11--20, 2016.

\bibitem{miao2023self}
Peihan Miao, Wei Su, Gaoang Wang, Xuewei Li, and Li Xi.
\newblock Self-paced multi-grained cross-modal interaction modeling for referring expression comprehension.
\newblock {\em IEEE Transactions on Image Processing}, 33:1497--1507, 2023.

\bibitem{nagaraja2016modeling}
Varun~K Nagaraja, Vlad~I Morariu, and Larry~S Davis.
\newblock Modeling context between objects for referring expression understanding.
\newblock In {\em ECCV}, pages 792--807, 2016.

\bibitem{plummerCITE2018}
Bryan~A. Plummer, Paige Kordas, M.~Hadi Kiapour, Shuai Zheng, Robinson Piramuthu, and Svetlana Lazebnik.
\newblock Conditional image-text embedding networks.
\newblock In {\em ECCV}, pages 249--264, 2018.

\bibitem{plummer2017flickr30k}
Bryan~A Plummer, Liwei Wang, Chris~M Cervantes, Juan~C Caicedo, Julia Hockenmaier, and Svetlana Lazebnik.
\newblock Flickr30k entities: Collecting region-to-phrase correspondences for richer image-to-sentence models.
\newblock {\em IJCV}, 123(1):74, 2017.

\bibitem{qi2020reverie}
Yuankai Qi, Qi Wu, Peter Anderson, Xin Wang, William~Yang Wang, Chunhua Shen, and Anton van~den Hengel.
\newblock Reverie: Remote embodied visual referring expression in real indoor environments.
\newblock In {\em Proceedings of the IEEE/CVF Conference on Computer Vision and Pattern Recognition}, pages 9982--9991, 2020.

\bibitem{qian2023multimodal}
Rui Qian, Yeqing Li, Zheng Xu, Ming-Hsuan Yang, Serge Belongie, and Yin Cui.
\newblock Multimodal open-vocabulary video classification via vision and language models, 2023.

\bibitem{radford2021learning}
Alec Radford, Jong~Wook Kim, Chris Hallacy, Aditya Ramesh, Gabriel Goh, Sandhini Agarwal, Girish Sastry, Amanda Askell, Pamela Mishkin, Jack Clark, et~al.
\newblock Learning transferable visual models from natural language supervision.
\newblock In {\em International conference on machine learning}, pages 8748--8763. PmLR, 2021.

\bibitem{rozenberszki2022language}
David Rozenberszki, Or Litany, and Angela Dai.
\newblock Language-grounded indoor 3d semantic segmentation in the wild.
\newblock In {\em European Conference on Computer Vision}, pages 125--141. Springer, 2022.

\bibitem{sadhu2019zero}
Arka Sadhu, Kan Chen, and Ram Nevatia.
\newblock Zero-shot grounding of objects from natural language queries.
\newblock In {\em ICCV}, pages 4694--4703, 2019.

\bibitem{shen2024groundvlp}
Haozhan Shen, Tiancheng Zhao, Mingwei Zhu, and Jianwei Yin.
\newblock Groundvlp: Harnessing zero-shot visual grounding from vision-language pre-training and open-vocabulary object detection.
\newblock In {\em Proceedings of the AAAI Conference on Artificial Intelligence}, volume~38, pages 4766--4775, 2024.

\bibitem{shi2023unpaired}
Hengcan Shi, Munawar Hayat, and Jianfei Cai.
\newblock Unpaired referring expression grounding via bidirectional cross-modal matching.
\newblock {\em Neurocomputing}, 518:39--49, 2023.

\bibitem{shi2022improving}
Zhan Shi, Yilin Shen, Hongxia Jin, and Xiaodan Zhu.
\newblock Improving zero-shot phrase grounding via reasoning on external knowledge and spatial relations.
\newblock In {\em Proceedings of the AAAI Conference on Artificial Intelligence}, volume~36, pages 2253--2261, 2022.

\bibitem{su2024scanformer}
Wei Su, Peihan Miao, Huanzhang Dou, and Xi Li.
\newblock Scanformer: Referring expression comprehension by iteratively scanning.
\newblock In {\em Proceedings of the IEEE/CVF Conference on Computer Vision and Pattern Recognition}, pages 13449--13458, 2024.

\bibitem{su2023language}
Wei Su, Peihan Miao, Huanzhang Dou, Gaoang Wang, Liang Qiao, Zheyang Li, and Xi Li.
\newblock Language adaptive weight generation for multi-task visual grounding.
\newblock In {\em Proceedings of the IEEE/CVF conference on computer vision and pattern recognition}, pages 10857--10866, 2023.

\bibitem{subramanian-etal-2022-reclip}
Sanjay Subramanian, Will Merrill, Trevor Darrell, Matt Gardner, Sameer Singh, and Anna Rohrbach.
\newblock Reclip: A strong zero-shot baseline for referring expression comprehension.
\newblock In {\em Proceedings of the 60th Annual Meeting of the Association for Computational Linguistics}, Dublin, Ireland, may 2022. Association for Computational Linguistics.

\bibitem{sun2023refteacher}
Jiamu Sun, Gen Luo, Yiyi Zhou, Xiaoshuai Sun, Guannan Jiang, Zhiyu Wang, and Rongrong Ji.
\newblock Refteacher: A strong baseline for semi-supervised referring expression comprehension.
\newblock In {\em Proceedings of the IEEE/CVF conference on computer vision and pattern recognition}, pages 19144--19154, 2023.

\bibitem{sun2021cycle}
Mingjie Sun, Jimin Xiao, Eng~Gee Lim, and Yao Zhao.
\newblock Cycle-free weakly referring expression grounding with self-paced learning.
\newblock {\em IEEE Transactions on Multimedia}, 25:1611--1621, 2021.

\bibitem{tang2020blockmix}
Hao Tang, Zechao Li, Zhimao Peng, and Jinhui Tang.
\newblock Blockmix: meta regularization and self-calibrated inference for metric-based meta-learning.
\newblock In {\em Proceedings of the 28th ACM international conference on multimedia}, pages 610--618, 2020.

\bibitem{tang2022learning}
Hao Tang, Chengcheng Yuan, Zechao Li, and Jinhui Tang.
\newblock Learning attention-guided pyramidal features for few-shot fine-grained recognition.
\newblock {\em Pattern Recognition}, 130:108792, 2022.

\bibitem{tang2023context}
Wei Tang, Liang Li, Xuejing Liu, Lu Jin, Jinhui Tang, and Zechao Li.
\newblock Context disentangling and prototype inheriting for robust visual grounding.
\newblock {\em IEEE Transactions on Pattern Analysis and Machine Intelligence}, 46(5):3213--3229, 2023.

\bibitem{vaswani2017attention}
Ashish Vaswani, Noam Shazeer, Niki Parmar, Jakob Uszkoreit, Llion Jones, Aidan~N Gomez, Lukasz Kaiser, and Illia Polosukhin.
\newblock Attention is all you need.
\newblock In {\em NeurIPS}, 2017.

\bibitem{wang2019phrase}
Josiah Wang and Lucia Specia.
\newblock Phrase localization without paired training examples.
\newblock In {\em Proceedings of the IEEE/CVF International Conference on Computer Vision}, pages 4663--4672, 2019.

\bibitem{wang2021improving}
Liwei Wang, Jing Huang, Yin Li, Kun Xu, Zhengyuan Yang, and Dong Yu.
\newblock Improving weakly supervised visual grounding by contrastive knowledge distillation.
\newblock In {\em Proceedings of the IEEE/CVF conference on computer vision and pattern recognition}, pages 14090--14100, 2021.

\bibitem{wang2019learning}
Liwei Wang, Yin Li, Jing Huang, and Svetlana Lazebnik.
\newblock Learning two-branch neural networks for image-text matching tasks.
\newblock {\em TPAMI}, 41:394--407, 2018.

\bibitem{wang2019neighbourhood}
Peng Wang, Qi Wu, Jiewei Cao, Chunhua Shen, Lianli Gao, and Anton van~den Hengel.
\newblock Neighbourhood watch: Referring expression comprehension via language-guided graph attention networks.
\newblock In {\em CVPR}, pages 1960--1968, 2019.

\bibitem{wang2022ofa}
Peng Wang, An Yang, Rui Men, Junyang Lin, Shuai Bai, Zhikang Li, Jianxin Ma, Chang Zhou, Jingren Zhou, and Hongxia Yang.
\newblock Ofa: Unifying architectures, tasks, and modalities through a simple sequence-to-sequence learning framework.
\newblock In {\em International conference on machine learning}, pages 23318--23340. PMLR, 2022.

\bibitem{wang2023image}
Wenhui Wang, Hangbo Bao, Li Dong, Johan Bjorck, Zhiliang Peng, Qiang Liu, Kriti Aggarwal, Owais~Khan Mohammed, Saksham Singhal, Subhojit Som, et~al.
\newblock Image as a foreign language: Beit pretraining for vision and vision-language tasks.
\newblock In {\em Proceedings of the IEEE/CVF Conference on Computer Vision and Pattern Recognition}, pages 19175--19186, 2023.

\bibitem{wang2021weakly}
Yuechen Wang, Jiajun Deng, Wengang Zhou, and Houqiang Li.
\newblock Weakly supervised temporal adjacent network for language grounding.
\newblock {\em IEEE Transactions on Multimedia}, 24:3276--3286, 2021.

\bibitem{xiao2024hivg}
Linhui Xiao, Xiaoshan Yang, Fang Peng, Yaowei Wang, and Changsheng Xu.
\newblock Hivg: Hierarchical multimodal fine-grained modulation for visual grounding.
\newblock In {\em Proceedings of the 32nd ACM International Conference on Multimedia}, pages 5460--5469, 2024.

\bibitem{xiao2023clip}
Linhui Xiao, Xiaoshan Yang, Fang Peng, Ming Yan, Yaowei Wang, and Changsheng Xu.
\newblock Clip-vg: Self-paced curriculum adapting of clip for visual grounding.
\newblock {\em IEEE Transactions on Multimedia}, 26:4334--4347, 2023.

\bibitem{xu2020attribute}
Wenjia Xu, Yongqin Xian, Jiuniu Wang, Bernt Schiele, and Zeynep Akata.
\newblock Attribute prototype network for zero-shot learning.
\newblock {\em Advances in Neural Information Processing Systems}, 33:21969--21980, 2020.

\bibitem{yang20243d}
Dejie Yang, Zhu Xu, Wentao Mo, Qingchao Chen, Siyuan Huang, and Yang Liu.
\newblock 3d vision and language pretraining with large-scale synthetic data.
\newblock In {\em Proceedings of the Thirty-Third International Joint Conference on Artificial Intelligence}, IJCAI '24, 2024.

\bibitem{yang2022improving}
Li Yang, Yan Xu, Chunfeng Yuan, Wei Liu, Bing Li, and Weiming Hu.
\newblock Improving visual grounding with visual-linguistic verification and iterative reasoning.
\newblock In {\em Proceedings of the IEEE/CVF Conference on Computer Vision and Pattern Recognition}, pages 9499--9508, 2022.

\bibitem{yang2019dynamic}
Sibei Yang, Guanbin Li, and Yizhou Yu.
\newblock Dynamic graph attention for referring expression comprehension.
\newblock In {\em ICCV}, pages 4644--4653, 2019.

\bibitem{yang2020improving}
Zhengyuan Yang, Tianlang Chen, Liwei Wang, and Jiebo Luo.
\newblock Improving one-stage visual grounding by recursive sub-query construction.
\newblock In {\em ECCV}, 2020.

\bibitem{yang2019fast}
Zhengyuan Yang, Boqing Gong, Liwei Wang, Wenbing Huang, Dong Yu, and Jiebo Luo.
\newblock A fast and accurate one-stage approach to visual grounding.
\newblock In {\em ICCV}, pages 4683--4693, 2019.

\bibitem{yao2024cpt}
Yuan Yao, Ao Zhang, Zhengyan Zhang, Zhiyuan Liu, Tat-Seng Chua, and Maosong Sun.
\newblock Cpt: Colorful prompt tuning for pre-trained vision-language models.
\newblock {\em AI Open}, 5:30--38, 2024.

\bibitem{yeh2018unsupervised}
Raymond~A Yeh, Minh~N Do, and Alexander~G Schwing.
\newblock Unsupervised textual grounding: Linking words to image concepts.
\newblock In {\em Proceedings of the IEEE Conference on Computer Vision and Pattern Recognition}, pages 6125--6134, 2018.

\bibitem{young2014image}
Peter Young, Alice Lai, Micah Hodosh, and Julia Hockenmaier.
\newblock From image descriptions to visual denotations: New similarity metrics for semantic inference over event descriptions.
\newblock {\em ACL}, 2:67--78, 2014.

\bibitem{yu2022coca}
Jiahui Yu, Zirui Wang, Vijay Vasudevan, Legg Yeung, Mojtaba Seyedhosseini, and Yonghui Wu.
\newblock Coca: Contrastive captioners are image-text foundation models.
\newblock {\em Transactions on Machine Learning Research}, 2022.

\bibitem{yu2018mattnet}
Licheng Yu, Zhe Lin, Xiaohui Shen, Jimei Yang, Xin Lu, Mohit Bansal, and Tamara~L Berg.
\newblock Mattnet: Modular attention network for referring expression comprehension.
\newblock In {\em CVPR}, pages 1307--1315, 2018.

\bibitem{yu2016modeling}
Licheng Yu, Patrick Poirson, Shan Yang, Alexander~C Berg, and Tamara~L Berg.
\newblock Modeling context in referring expressions.
\newblock In {\em ECCV}, pages 69--85, 2016.

\bibitem{yu2017joint}
Licheng Yu, Hao Tan, Mohit Bansal, and Tamara~L Berg.
\newblock A joint speaker-listener-reinforcer model for referring expressions.
\newblock In {\em CVPR}, pages 7282--7290, 2017.

\bibitem{yu2018rethinking}
Zhou Yu, Jun Yu, Chenchao Xiang, Zhou Zhao, Qi Tian, and Dacheng Tao.
\newblock Rethinking diversified and discriminative proposal generation for visual grounding.
\newblock In {\em IJCAI}, 2018.

\bibitem{zhan2023object}
Zhaohuan Zhan, Liang Lin, and Guang Tan.
\newblock Object-aware navigation for remote embodied visual referring expression.
\newblock {\em Neurocomputing}, 515:68--78, 2023.

\bibitem{zhang2018grounding}
Hanwang Zhang, Yulei Niu, and Shih-Fu Chang.
\newblock Grounding referring expressions in images by variational context.
\newblock In {\em CVPR}, pages 4158--4166, 2018.

\bibitem{zheng2024resvg}
Minghang Zheng, Jiahua Zhang, Qingchao Chen, Yuxin Peng, and Yang Liu.
\newblock Resvg: Enhancing relation and semantic understanding in multiple instances for visual grounding.
\newblock In {\em Proceedings of the 32nd ACM International Conference on Multimedia}, pages 1187--1196, 2024.

\bibitem{zhu2022seqtr}
Chaoyang Zhu, Yiyi Zhou, Yunhang Shen, Gen Luo, Xingjia Pan, Mingbao Lin, Chao Chen, Liujuan Cao, Xiaoshuai Sun, and Rongrong Ji.
\newblock Seqtr: A simple yet universal network for visual grounding.
\newblock In {\em Computer Vision--ECCV 2022: 17th European Conference, Tel Aviv, Israel, October 23--27, 2022, Proceedings, Part XXXV}, pages 598--615. Springer, 2022.

\bibitem{zhu2021utilizing}
Haidong Zhu, Arka Sadhu, Zhaoheng Zheng, and Ram Nevatia.
\newblock Utilizing every image object for semi-supervised phrase grounding.
\newblock In {\em Proceedings of the IEEE/CVF Winter Conference on Applications of Computer Vision}, pages 2210--2219, 2021.

\bibitem{zhuang2018parallel}
Bohan Zhuang, Qi Wu, Chunhua Shen, Ian Reid, and Anton van~den Hengel.
\newblock Parallel attention: A unified framework for visual object discovery through dialogs and queries.
\newblock In {\em CVPR}, pages 4252--4261, 2018.

\bibitem{ziegel2003elements}
Eric~R Ziegel.
\newblock The elements of statistical learning, 2003.

\end{thebibliography}
}
\appendix
\section{Preliminary}
\label{sec:Preliminary}
Before delving into the specifics of the PAML model, we will first provide a brief introduction to three distinct types of attention mechanisms~\cite{vaswani2017attention}.

\subsection{Self-Attention Mechanism}
\label{sec:Self-Attention} Mechanism
The Self-Attention mechanism is employed to capture relationships among elements within a single sequence. Given an input sequence $X = (x_1, x_2, \dots, x_n)$, where $x_i \in \mathbb{R}^{d}$, the computation of Self-Attention proceeds as follows:

\begin{equation}
	\label{eq:selfattentionwieght}
	Q=XW_Q, K=XW_k, V=XW_V 
\end{equation}
Here $W_Q,W_k,W_v \in \mathbb{R}^{d\times d_k}$
\begin{equation}
	\label{eq:selfattentioncompute}
	Attention(Q,K,V)=softmax(\frac{QK^T}{\sqrt{d_k}})V 
\end{equation}
In Eq.~\eqref{eq:selfattentioncompute}, $\frac{QK^T}{\sqrt{d_k}}$ represents the scaled dot-product attention, and $d_k$ denotes the dimensionality of the key vectors.
Multi-head self-attention mechanism is a variant of single-head self-attention mechanism. While it is structurally similar to Eq.~\eqref{eq:selfattentioncompute} in form, each attention head can be regarded as learning relationships between elements in the sequence from different "perspectives." This introduction enhances the model's ability to generalize to unseen data. In the subsequent sections, we will use $MSA(\cdot)$ to denote the Multi-head Self-Attention mechanism operation.

\subsection{Cross-Attention Mechanism}
\label{sec:Cross-Attention Mechanism}
The Cross-Attention mechanism is designed to capture interactions between different sequences. Given two input sequences from distinct modalities. $X=(x_1, x_2, \dots, x_n)$ and $Y=(y_1, y_2, \dots, y_n)$, where $x_i \in \mathbb{R}^{d_x}$ and $y_j \in \mathbb{R}^{d_y}$, the computation of Cross-Attention proceeds as follows:
\begin{equation}
	\label{eq:crossattentionwieght}
	Q=XW_Q, K=YW_k, V=YW_V 
\end{equation}
Here, $W_Q \in \mathbb{R}^{d_x\times d_k}$,and $W_k,W_V \in \mathbb{R}^{d_y\times d_k}$
\begin{equation}
	\label{eq:crossattentioncompute}
	\text{Cross-attention}(Q,K,V)=softmax(\frac{QK^T}{\sqrt{d_k}})V
\end{equation}
Similarly, we use $MCA(\cdot)$ to denote the Multi-head Cross-ateention mechanism operation.

\subsection{Attention mechanism+Relative Positional Encoding}
\label{sec:Attention mechanism+Relative Positional Encoding}
This is an enhanced attention mechanism that combines attention mechanism with relative positional encoding, capable of simultaneously capturing both the semantic relationships and the relative positional relationships among elements in sequences.
Given three different sequences, $X=(x_1, x_2, \dots, x_n)$, $Y=(y_1, y_2, \dots, y_n)$ and $Z=(z_1, z_2, \dots, z_n)$.
where $x_i \in \mathbb{R}^{d_x}$, $y_j \in \mathbb{R}^{d_y}$ and $z_t \in \mathbb{R}^{d_z}$  the computation of Cross-Attention PRE proceeds as follows:
\begin{equation}
	\label{eq:crossattentionRPEwieght}
	Q=XW_Q, K=YW_k, V=ZW_V 
\end{equation}
Here, $W_Q \in \mathbb{R}^{d_x\times d_k}$, $W_k\in \mathbb{R}^{d_y\times d_k}$ and $W_V \in \mathbb{R}^{d_z\times d_k}$
\begin{equation}
	\label{eq:crossattentionRPEcompute}
	\text{AttentionRPE}(Q,K,V)=softmax(\frac{QK^T+R}{\sqrt{d_k}})V
\end{equation}
$R$ represents the relative positional encoding, which is utilized to model the relative positional relationships between elements. The specific computation method is as follows:

Let us define the following quantities:
\begin{itemize}
    \item Batch size $B$, number of attention heads $h$, embedding dimension $dim$, and image height and width dimension H, W
    \item Input mask $\bm{M} \in \{0,1\}^{B \times HW}$ with $\bm{M}_{b,i} = \text{key\_padding\_mask}[b,i]$
    \item Image mask $\bm{I} \in \{0,1\}^{B \times H \times W} = \neg \bm{M}$ (reshaped)
    \item Query matrix $\bm{Q} \in \mathbb{R}^{(B \cdot h) \times (dim/h) \times HW}$
    \item Key weight matrix $\bm{W}_k \in \mathbb{R}^{dim \times dim}$
\end{itemize}
First compute the cumulative positions along rows and columns:

\begin{equation}
	\label{eq:cumulative positions}
	\begin{aligned}
    	\bm{yy}_{b,i,j} &= \sum_{k=1}^j \bm{I}_{b,i,k} \quad \forall b \in [1,B], i \in [1,H], j \in [1,H] \\
    	\bm{xx}_{b,i,j} &= \sum_{k=1}^i \bm{I}_{b,k,j} \quad \forall b \in [1,B], i \in [1,W], j \in [1,W]
	\end{aligned}
\end{equation}

Flatten these to $\bm{yy}, \bm{xx} \in \mathbb{R}^{B \times HW}$, then compute relative position differences:

\begin{equation}
	\label{eq:relative position differences}
	\begin{aligned}
    	\Delta\bm{yy} &= \bm{yy}_{:, :, \text{None}} - \bm{yy}_{:, \text{None}, :} \in \mathbb{R}^{B \times HW \times HW} \\
    	\Delta\bm{xx} &= \bm{xx}_{:, :, \text{None}} - \bm{xx}_{:, \text{None}, :} \in \mathbb{R}^{B \times HW \times HW}
	\end{aligned}	
\end{equation}

Define position mbeddeing matrices (either learnable or fixed):

\begin{equation}
	\label{eq:position mbeddeing matrices}
	\begin{aligned}
    	\bm{P}_y &\in \mathbb{R}^{y\_range \times (dim/2)} \\
    	\bm{P}_x &\in \mathbb{R}^{x\_range \times (dim/2)}
	\end{aligned}	
\end{equation}

Project through key weights:

\begin{equation}
	\label{eq:key weights}
	\begin{aligned}
    	\bm{K}_y &= \bm{P}_y \bm{W}_k[:dim/2,:]^T \in \mathbb{R}^{y\_range \times dim} \\
    	\bm{K}_x &= \bm{P}_x \bm{W}_k[dim/2:,:]^T \in \mathbb{R}^{x\_range \times dim}
	\end{aligned}	
\end{equation}

Reshape and prepare for multi-head attention:

\begin{equation}
	\label{eq:Reshape for MHA}
	\begin{aligned}
    	\bm{K}'_y &\in \mathbb{R}^{(B \cdot h) \times (dim/h) \times y\_range} \\
    	\bm{K}'_x &\in \mathbb{R}^{(B \cdot h) \times (dim/h) \times x\_range}
	\end{aligned}
\end{equation}

Compute raw attention scores:

\begin{equation}
	\label{eq:Reshape for MHA}
	\begin{aligned}
    	\bm{A}^{raw}_y &= \bm{Q} \bm{K}'_y{}^T \in \mathbb{R}^{(B \cdot h) \times HW \times y\_range} \\
    	\bm{A}^{raw}_x &= \bm{Q} \bm{K}'_x{}^T \in \mathbb{R}^{(B \cdot h) \times HW \times x\_range}
	\end{aligned}
\end{equation}

Reshape and compute position indices:

\begin{equation}
	\label{eq:position indices}
	\begin{aligned}
    	\bm{I}_{yy} &= \Delta\bm{yy}_{:, \text{None}, :, :} + \text{pos\_index\_offset} \in \mathbb{Z}^{B \times h \times HW \times HW} \\
    	\bm{I}_{xx} &= \Delta\bm{xx}_{:, \text{None}, :, :} + \text{pos\_index\_offset} \in \mathbb{Z}^{B \times h \times HW \times HW}
	\end{aligned}
\end{equation}

Gather final position attention scores:

\begin{equation}
	\label{eq:position attention scores}
	\begin{aligned}
    	\bm{A}_y[b,h,i,j] &= \bm{A}^{raw}_y[b,h,i, \bm{I}_{yy}[b,h,i,j]] \\
    	\bm{A}_x[b,h,i,j] &= \bm{A}^{raw}_x[b,h,i, \bm{I}_{xx}[b,h,i,j]]
	\end{aligned}
\end{equation}

Combine the position attention scores:

\begin{equation}
	\label{eq:position attention scores}
	\begin{aligned}
    	R = (\bm{A}_y + \bm{A}_x) \in \mathbb{R}^{(B \cdot h) \times HW \times HW}
	\end{aligned}
\end{equation}

\end{document}